\documentclass[11pt]{article}

\usepackage[in]{fullpage}

\usepackage{amsfonts,amscd,float}
\usepackage{array,booktabs,paralist}
\usepackage{graphicx,verbatim}
\usepackage[rflt]{floatflt}
\usepackage{times,graphicx}
\usepackage{amsmath,amssymb,amsopn,algorithm,algorithmic,float,bbm,bm,enumerate,color,multirow}
\usepackage{setspace}
\usepackage{rotating}
\usepackage{array}	
\usepackage{hyperref}
\usepackage{url}
\usepackage{wrapfig}
\usepackage{subfigure}

\usepackage{epstopdf}
\usepackage{nameref}
\usepackage{zref-xr}
\usepackage{sidecap}
\newcommand{\beq}{\begin{equation}}
\newcommand{\eeq}{\end{equation}}

\newcommand{\hth}{\hat{\theta}}

\newcommand\R{\mathbb{R}}

\newcommand{\cC}{{\cal C}}
\newcommand{\cE}{{\cal E}}
\newcommand{\cH}{{\cal H}}

\newcommand{\cN}{{\cal N}}

\newcommand{\cS}{{\cal S}}

\newcommand{\cA}{{\cal A}}

\newcommand{\vertiii}[1]{{\left\vert\kern-0.25ex\left\vert\kern-0.25ex\left\vert #1
    \right\vert\kern-0.25ex\right\vert\kern-0.25ex\right\vert}}

\newcommand{\Th}{\boldsymbol{\theta}}

\newcommand{\oomega}{\boldsymbol{\omega}}

\newcommand{\myref}[1]{(\ref{#1})}

\DeclareMathOperator{\argmin}{argmin}

\DeclareMathOperator{\conv}{conv}

\DeclareMathOperator{\supp}{supp}

\DeclareMathOperator{\cone}{cone}

\newcommand{\proof}{\noindent{\itshape Proof:}\hspace*{1em}}

\newcommand{\qed}{\nolinebreak[1]~~~\hspace*{\fill} \rule{5pt}{5pt}\vspace*{\parskip}\vspace*{1ex}}

\newtheorem{theorem}{Theorem}
\newtheorem{proposition}[theorem]{Proposition}
\newtheorem{lemma}{Lemma}

\newtheorem{example}{Example}

\newcommand {\commentout}[1] {}

\def\ints{{{\rm Z} \kern -.35em {\rm Z} }}  
\def\smallints{{{\rm Z} \kern -.3em {\rm Z} }}  
\def\pints{{{\rm I} \kern -.15em {\rm N} }}      
\newcommand{\reals}{\mathbb R}

\def\cplx{{{\rm I} \kern -.45em {\rm C} }}       
\def\l2{\rm {\mathcal L}^{2}(\reals)}            

\newcommand{\be}{\begin{eqnarray}}
\newcommand{\ee}{\end{eqnarray}}
\newcommand{\bea}{\begin{eqnarray}}
\newcommand{\eea}{\end{eqnarray}}
\newcommand{\beaa}{\begin{eqnarray*}}
\newcommand{\eeaa}{\end{eqnarray*}}
\newcommand{\bnad}{\begin{nad}}
\newcommand{\enad}{\end{nad}}

\newcommand{\lip}{\langle}
\newcommand{\rip}{\rangle}

\newcommand{\sign}{{\mbox{\rm sign}}}

\renewcommand{\supp}{{\mbox{\rm supp}}}

\renewcommand{\overline}{\bar}

\title{High Dimensional Structured Superposition Models}
\author{Qilong Gu \\
Dept of Computer Science \& Engineering\\
University of Minnesota, Twin Cities\\
guxxx396@cs.umn.edu
\and
Arindam Banerjee\\
Dept of Computer Science \& Engineering\\
University of Minnesota, Twin Cities\\
banerjee@cs.umn.edu}

\begin{document}

\maketitle


\begin{abstract}

High dimensional superposition models characterize observations using parameters which can be written as a sum of multiple component parameters, each with its own structure, e.g., sum of low rank and sparse matrices, sum of sparse and rotated sparse vectors, etc. In this paper, we consider general superposition models which allow sum of any number of component parameters, and each component structure can be characterized by any norm. We present a simple estimator for such models, give a geometric condition under which the components can be accurately estimated, characterize sample complexity of the estimator, and give high probability non-asymptotic bounds on the componentwise estimation error. We use tools from empirical processes and generic chaining for the statistical analysis, and our results, which substantially generalize prior work on superposition models, are in terms of Gaussian widths of suitable sets.

%
\end{abstract}

\section{Introduction}
\label{sec:intro}
For high-dimensional structured estimation problems \cite{buva11, tibs96}, considerable advances have been made in accurately estimating a sparse or structured parameter $\theta \in \R^p$ even when the sample size $n$ is far smaller than the ambient dimensionality of $\theta^*$, i.e., $n \ll p$. Instead of a single structure, such as sparsity or low rank, recent years have seen interest in parameter estimation when the parameter $\theta^*$ is a {\em superposition} or {\em sum of multiple different structures}, i.e., $\theta = \sum_{i=1}^k \theta_i^*$, where $\theta_1^*$ may be sparse, $\theta_2^*$ may be low rank, and so on \cite{agnw12,clmw11,chpw12,cspw11,foma14,hskz11,mctr13,mcco13,wgmm12,yara12}.

In this paper, we substantially generalize the non-asymptotic estimation error analysis for such superposition models such that (i) the parameter $\theta$ can be the superposition of {\em any number of component parameters} $\theta_i^*$, and (ii) the structure in each $\theta_i^*$ can be captured by {\em any suitable norm} $R_i(\theta_i^*)$. We will analyze the following linear measurement based superposition model
\begin{equation}\label{modsp}
  y = X\sum_{i=1}^k \theta_i^* + \omega~,
\end{equation}
where $X \in \R^{n \times p}$ is a random sub-Gaussian design or compressive matrix, $k$ is the number of components, $\theta_i^*$ is one component of the unknown parameters, $y \in \R^n$ is the response vector, and $\omega \in \R^n$ is random noise independent of $X$. The structure in each component $\theta_i^*$ is captured by any suitable norm $R_i(\cdot)$, such that $R_i(\theta_i^*)$ has a small value, e.g., sparsity captured by $\| \theta_i^* \|_1$, low-rank (for matrix $\theta_i^*$) captured by the nuclear norm $\| \theta_i^* \|_*$, etc. Popular models such as Morphological Component Analysis (MCA) \cite{dohu01} and Robust PCA \cite{clmw11,cspw11} can be viewed as a special cases of this framework (see Section~\ref{sec:examples}).

The superposition estimation problem can be posed as follows: Given $(y,X)$ generated following \myref{modsp}, estimate component parameters $\{ \hat{\theta}_i \}$ such that all the component-wise estimation errors $\Delta_i = \hat{\theta}_i - \theta_i^*$, where $\theta^*_i$ is the population mean, are small. Ideally, we want to obtain high-probability non-asymptotic bounds on the total componentwise error measured as $\sum_{i=1}^k \| \hat{\theta}_i - \theta_i^* \|_2$, with the bound improving (getting smaller) with increase in the number $n$ of samples.

We propose the following estimator for the superposition model in (\ref{modsp}):
\beq
    \min_{\{\theta_1,\ldots,\theta_k\}} ~\left\|y- X \sum_{i=1}^k \theta_i
    \right\|_2^2 \qquad \text{s.t.}~\quad R_i(\theta_i) \leq \alpha_i~, \quad
    i = 1, \ldots, k~,
    \label{eq:est1}
\eeq
where $\alpha_i$ are suitable constants. In this paper, we focus on the case where $\alpha_i = R_i(\theta_i^*)$, 
noting that recent advances~\cite{oyrs15} can be used to extend our results to more general settings.


The superposition estimator in \myref{eq:est1} succeeds if a certain geometric condition, which we call {\em structural coherence} (SC), is satisfied by certain sets (cones) associated with the component norms $R_i(\cdot)$. Since the estimate $\hat{\theta}_i = \theta_i^* + \Delta_i$ is in the feasible set of the optimization problem \myref{eq:est1}, the error vector $\Delta_i$ satisfies the constraint $R_i(\theta^*_i + \Delta_i) \leq \alpha_i$ where $\alpha_i = R_i(\theta^*_i)$. The SC condition of is a geometric relationship between the corresponding error cones $\cC_i = \cone\{ \Delta_i | R_i(\theta_i^* + \Delta_i) \leq R_i(\theta_i^*)\}$ (see Section~\ref{sec:geometry}).

If SC is satisfied, then we can show that the sum of componentwise estimation error can be bounded with high probability, and the bound takes the form:
\beq
\sum_{i=1}^k \| \hat{\theta}_i - \theta_i^* \|_2 \leq c \frac{ \max_{i} w(\cC_i \cap B_p) + \sqrt{\log k} }{\sqrt{n}}~,
\label{eq:bnd0}
\eeq
where $n$ is the sample size, $k$ is the number of components, and $w(\cC_i \cap B_p)$ is the Gaussian width \cite{bcfs14,crpw12,vers14} of the intersection of the error cone $\cC_i$ with the unit Euclidean ball $B_p \subseteq \R^p$. Interestingly, the estimation error converges at the rate of $\frac{1}{\sqrt{n}}$, similar to the case of single parameter estimators \cite{nrwy12,bcfs14}, and depends only logarithmically on the number of components $k$. Further, while dependency of the error on Gaussian width of the error has been shown in recent results involving a single parameter \cite{bcfs14,vers14}, the bound in \myref{eq:bnd0} depends on the maximum of the Gaussian width of individual error cones, not their sum. The analysis thus gives a general way to construct estimators for superposition problems along with high-probability non-asymptotic upper bounds on the sum of componentwise errors.
To show the generality of our work, we provide a detailed review and comparison with related work in Appendix \ref{sec:related2}.

{\bf Notation}: In this paper, we use $\|.\|$ to denote vector norm, and $\vertiii{.}$ to denote operator norm. For example, $\|.\|_2$ is the Euclidean norm for a vector or matrix, and $\vertiii{.}_*$ is the nuclear norm of a matrix. We denote $\cone\{\cE\}$ as the smallest closed cone that contains a given set $\cE$. We denote $\langle ., .\rangle$ as the inner product.

The rest of this paper is organized as follows: We start with an optimization algorithm in Section \ref{sec:opt} and a deterministic estimation error bound in Section~\ref{sec:reco}, while laying down the key geometric and statistical quantities involved in the analysis.
In Section \ref{sec:geometry}, we discuss the geometry of the structural coherence (SC) condition, 
and show that the geometric SC condition implies statistical restricted eigenvalue (RE) condition. 
In Section~\ref{sec:desnoise}, we develop the main error bound on the sum of componentwise errors which hold with high probability for sub-Gaussian designs and noise.
In Section \ref{sec:related}, we compare an estimator using ``infimal convolution''\cite{rock70} of norms with our estimator \myref{eq:est1} for the noiseless case. We discuss related work in Section \ref{sec:related2}. We apply our error bound to practical problems in Section \ref{sec:examples}, present experimental results in Section \ref{sec:expr}, and conclude in Section \ref{sec:conc}.
The proofs of all technical results are in the Appendix. 

\section{Error Structure and Recovery Guarantees}
\label{sec:reco}



In this section, we start with some basic results and, under suitable assumptions, provide a deterministic bound for the componentwise estimation error in superposition models. Subsequently, we will show that the assumptions made here hold with high probability as long as a purely geometric non-probabilistic condition characterized by structural coherence (SC) is satisfied.

Let $\{ \hat{\theta}_i\}$ be a solution to the superposition estimation problem in \myref{eq:est1}, $\{\theta^*_i\}$ be the optimal (population) parameters involved in the true data generation process. Let $\Delta_i = \hat{\theta}_i - \theta_i^*$ be the error vector for component $i$ of the superposition. Our goal is to provide a preliminary understanding of the structure of error sets where $\Delta_i$ live, identify conditions under which a bound on the total componentwise error $\sum_{i=1}^k\|\hat{\theta}_i-\theta_i^*\|_2$ will hold, and provide a preliminary version of such a bound, which will be subsequently refined to the form in \myref{eq:bnd0} in Section~\ref{sec:desnoise}.  
Since $\hat{\theta}_i = \theta_i^* + \Delta_i$ lies in the feasible set of \myref{eq:est1}, as discussed in Section~\ref{sec:intro}, the error vectors $\Delta_i$ will lie in the error sets $\cE_i = \{ \Delta_i \in \R^p | R_i(\theta_i^*+\Delta_i) \leq R_i(\theta_i^*) \}$ respectively. For the analysis, we will be focusing on the cone of such error sets, given by
\beq
\cC_i = \cone \{ \Delta_i \in \R^p | R_i(\theta_i^*+\Delta_i) \leq R_i(\theta_i^*) \}~.
\eeq
Let $\theta^* = \sum_{i=1}^k \theta_i^*$, $\hat{\theta} = \sum_{i=1}^k \hat{\theta}_i$, and $\Delta = \sum_{i=1}^k \Delta_i$, so that $\Delta = \hat{\theta} - \theta^*$. From the optimality of $\hat{\theta}$ as a solution to \myref{eq:est1}, we have
\beq
\| y - X \hat{\theta} \|^2 \leq \| y -X \theta^* \|^2 ~ \Rightarrow ~\| X \Delta \|^2 \leq 2 \omega^T X \Delta~,
\label{eq:basic-ineq}
\eeq
using $\hat{\theta} = \theta^* + \Delta$ and $y = X \theta^* + \omega$. 
In order to establish recovery guarantees, under suitable assumptions we construct a lower bound to $\| X \Delta \|^2$, the left hand side of \myref{eq:basic-ineq}. The lower bound is a generalized form of the 
{\em restricted eigenvalue} (RE) condition studied in the literature \cite{birt09,buva11,rawy10}.
We also construct an upper bound to $\omega^T X \Delta$, the right hand side of \myref{eq:basic-ineq}, which needs to carefully analyze the noise-design (ND) interaction, i.e.,  between the noise $\omega$ and the design $X$.

We start by assuming that a generalized form of RE condition is satisfied by the superposition of errors: there exists a constant $\kappa > 0$ such that for all $\Delta_i \in \cC_i, i=1,2,\ldots,k$:
\begin{equation}
\hspace*{-3cm}\text{(RE)} \qquad \qquad \qquad \frac{1}{\sqrt{n}} \left\|X \sum_{i=1}^k \Delta_i\right\|_2 \geq \kappa \sum_{i=1}^k \|\Delta_i\|_2~.
\label{eqre}
\end{equation}
The above RE condition considers the following set:
\begin{equation}
  \cH = \left\{ \sum_{i=1}^k \Delta_i : \Delta_i \in \cC_i, \sum_{i=1}^k \| \Delta_i \|_2
  =	 1\right\}~.
  \label{eq:setC}
\end{equation}
which involves all the $k$ error cones, and the lower bound is over the sum of norms of the component wise errors.
If $k=1$, the RE condition in \myref{eqre} above simplifies to the widely studied RE condition in the current literature on Lasso-type and Dantzig-type estimators~\cite{birt09,rawy10,bcfs14} where only one error cone is involved. If we set all components but $\Delta_i$ to zero, then \myref{eqre} becomes the RE condition only for component $i$. We also note that the general RE condition as explicitly stated in \myref{eqre} has been implicitly used in \cite{agnw12} and \cite{yara12}.
For subsequent analysis, we introduce the set $\overline{\cH}$ defined as
\beq
\overline{\cH} = \left\{ \sum_{i=1}^k \Delta_i : \Delta_i \in \cC_i, \sum_{i=1}^k \| \Delta_i \|_2
\leq	 1\right\}.
\label{eq:setBC}
\eeq
noting that $\cH \subset \overline{\cH}$.


The general RE condition in \myref{eqre} depends on the random design matrix $X$, and is hence an inequality which will hold with certain probability depending on $X$ and the set $\cH$. For superposition problems, the probabilistic RE condition as in \myref{eqre} is intimately related to the following deterministic {\em structural coherence} (SC) condition on the interaction of the different component cones $\cC_i$, without any explicit reference to the random design matrix $X$: there is a constant $\rho > 0$ such that
for all $\Delta_i\in \cC_i, i=1,\ldots,k$,
\begin{equation}
\hspace*{-3cm} \text{(SC)} \qquad \qquad \qquad 
\left\| \sum_{i=1}^k \Delta_i \right\|_2 \geq \rho \sum_{i=1}^k \| \Delta_i \|_2~.
\label{cond2}
\end{equation}
If $k=1$, the SC condition is trivially satisfied with $\rho=1$. Since most existing literature on high-dimensional structured models focus on the $k=1$ setting~\cite{birt09,rawy10,bcfs14}, there was no reason to study the SC condition carefully.
For $k>1$, the SC condition \myref{cond2} implies a non-trivial relationship among the component cones. In particular, if the SC condition is true, then the sum $\sum_{i=1}^k \Delta_i$ being zero implies that each component $\Delta_i$ must also be zero. As presented in \myref{cond2}, the SC condition comes across as an algebraic condition. In Section~\ref{sec:geometry}, we present a geometric characterization of the SC condition \cite{mctr13}, and illustrate that the condition is both necessary and sufficient for accurate recovery of each component. In Section~\ref{sec:re}, we show that for sub-Gaussian design matrices $X$, the SC condition in \myref{cond2} in fact implies that the RE condition in \myref{eqre} will hold with high probability, after the number of samples crosses a certain sample complexity, which depends on the Gaussian width of the component cones. For now, we assume the RE condition in \myref{eqre} to hold, and proceed with the error bound analysis.

To establish recovery guarantee, following \myref{eq:basic-ineq}, we need an upper bound on the interaction between noise $\oomega$ and design $X$ \cite{bcfs14,mend14}. In particular, we consider the {\em noise-design} (ND) interaction
\beq
\hspace*{-3cm} \text{(ND)} \qquad \qquad 
s_n(\gamma) = \inf_{s>0}\left\{s:\sup_{u\in s\cH} \frac{1}{\sqrt{n}}\omega^T  X u \leq \gamma s^2 \sqrt{n}\right\}~,
\label{eq:nd1}
\eeq
where $\gamma > 0$ is a constant, and $s\cH$ is the scaled version of $\cH$ where the scaling factor is $s > 0$. Here, $s_n(\gamma)$ denotes the minimal scaling needed on $\cH$ such that one obtains a uniform bound over $\Delta \in s \cH$ of the form: $\frac{1}{n} \omega^T X \Delta \leq \gamma s_n^2(\gamma)$. Then, from the basic inequality in \myref{eq:basic-ineq}, with the bounds implied by the RE condition and the ND interaction, we have
\beq
\frac{1}{\sqrt{n}} \| X \Delta \|_2 \leq \frac{1}{\sqrt{n}} \sqrt{\omega^T X \Delta} \quad \Rightarrow \quad \kappa \sum_{i=1}^k \| \Delta_i \|_2 \leq \sqrt{\gamma} s_n(\gamma)~,
\eeq
which implies a bound on the component-wise error. The main deterministic bound below states the result formally:
\begin{theorem}[Deterministic bound]\label{thdet}
Assume that the RE condition in \myref{eqre} is satisfied in $\cH$ with parameter $\kappa$. Then, if $\kappa^2 > \gamma$, we have
$
  \sum_{i=1}^k \|\Delta_i\|_2 \leqslant 2s_n(\gamma)
$.
\end{theorem}
The above bound is deterministic and holds only when the RE condition in \myref{eqre} is satisfied with constant $\kappa$ such that $\kappa^2 > \gamma$. In the sequel, we first give a geometric characterization of the SC  condition in Section~\ref{sec:geometry}, and show that the SC condition implies the RE condition with high probability in Section~\ref{sec:re}. Further, we give a high probability characterization of $s_n(\gamma)$ based on the noise $\oomega$ and design $X$ in terms of the Gaussian widths of the component cones, and also illustrate how one can choose $\gamma$ in Section~\ref{sec:desnoise}. With these characterizations, we will obtain the desired component-wise error bound of the form \myref{eq:bnd0}.

\section{Geometry of Structural Coherence}
\label{sec:geometry}
In this section, we give a geometric characterization of the structural coherence (SC) condition in \myref{cond2}.
We start with the simplest case of two vectors $x,y$. If they are not reflections of each other, i.e., $x \neq -y$, then the following relationship holds:




\begin{proposition}\label{lemta}
  If there exists a $\delta < 1$ such that $-\langle x, y\rangle \leq \delta \| x \|_2 \| y \|_2$, then
  \begin{equation}\label{eqtr}
      \| x + y \|_2 \geq \sqrt{\frac{1 - \delta}{2}} (\| x \|_2 + \| y
     \|_2)~.
  \end{equation}
\end{proposition}

\begin{figure}[t]
	\centering
	\includegraphics[width=0.8\textwidth]{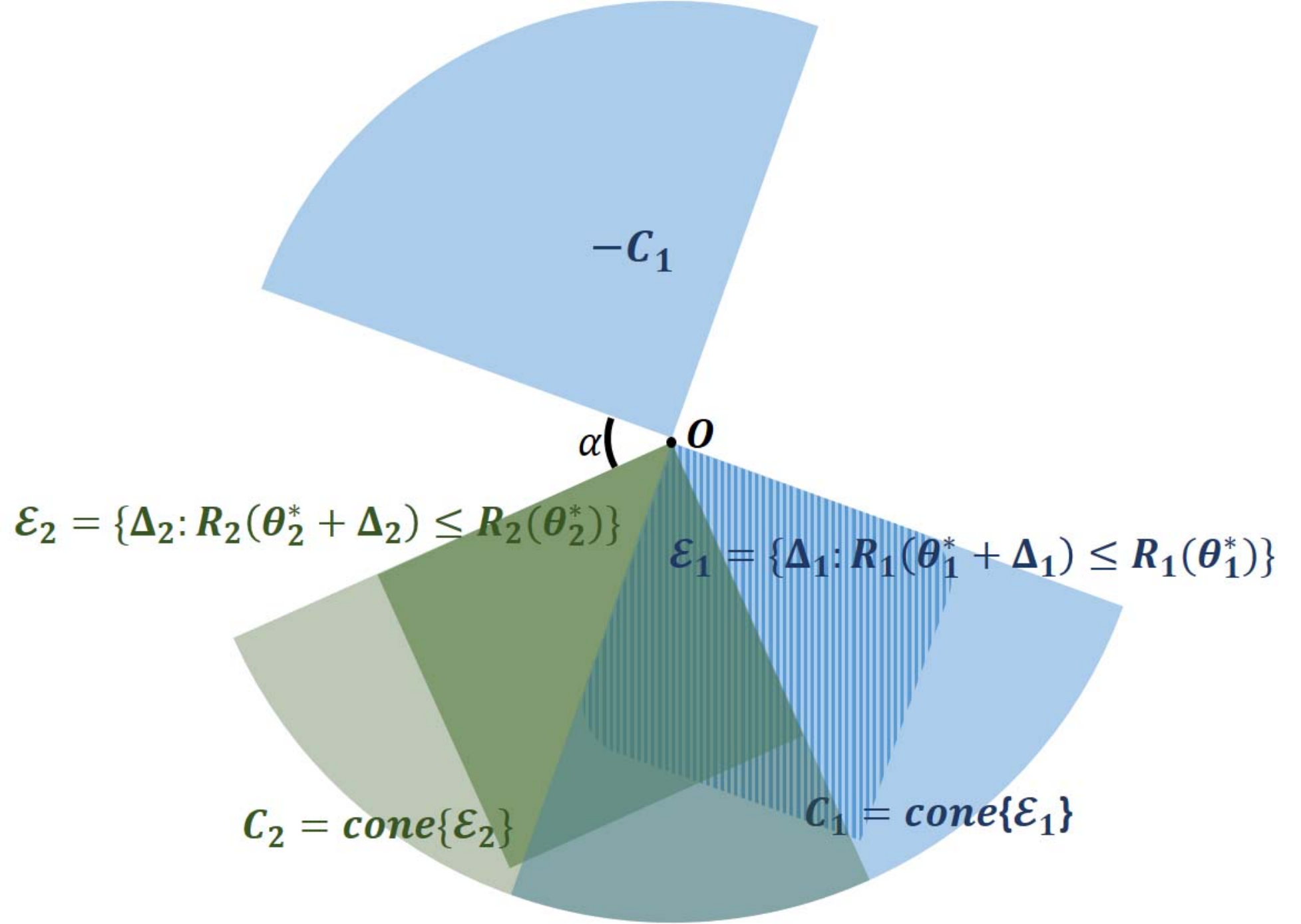}
	\caption{Geometry of SC condition when $k=2$. The error sets $\cE_1$ and $\cE_2$ are respectively shown as
blue and green squares, and the corresponding error cones are $\cC_1$ and $\cC_2$ respectively. $-\cC_1$ is the reflection of error cone $\cC_1$. If $-\cC_1$ and $\cC_2$ do not share a ray, i.e., the angle $\alpha$ between the cones is larger than $0$, then $\delta_0<1$, and the SC condition will hold.}
\end{figure}

Next, we generalize the condition of Proposition~\ref{lemta} to vectors in two different cones $\cC_1$ and
$\cC_2$. Given the cones, define
\beq
\delta_0 = \underset{x\in\cC_1\cap\cS^{p-1}, y\in\cC_2\cap\cS^{p-1}}{\sup} -\lip x,y \rip~.
\eeq
By construction, $-\lip x, y\rip \leq \delta_0\|x\|_2\|y\|_2$ for all $x\in\cC_1$ and $y\in\cC_2$.
If $\delta_0 < 1$, then $\myref{eqtr}$ continues to hold for all $x\in\cC_1$ and $y\in\cC_2$ with constant $\sqrt{(1-\delta_0)/2} > 0$. Note that this corresponds to the SC condition with $k=2$ and $\rho = \sqrt{(1-\delta_0)/2}$. We can interpret this geometrically as follows: first reflect cone $\cC_1$ to get $-\cC_1$, then $\delta$ is the cosine of the minimum angle between $-\cC_1$ and $\cC_2$. If $\delta_0 = 1$,
then $-\cC_1$ and $\cC_2$ share a ray, and structural coherence does not hold. Otherwise, $\delta_0 < 1$, implying $-\cC_1 \cap \cC_2 = \{0\}$, i.e., the two cones intersect only at the origin, and structural coherence holds.

For the general case involving $k$ cones, denote
\beq
\delta_i = \sup_{u \in -\cC_i \cap S_{p - 1}, v \in \sum_{j \neq i} \cC_j \cap
   S_{p - 1}} \langle u, v \rangle~.
\label{eq:deltai}
\eeq
In recent work, \cite{mctr13} concluded that if $\delta_i<1$ for each $i=1,\ldots,k$ then $-\cC_i$ and $\sum_{j\neq i}\cC_j$ does not share a ray, and the original signal can be recovered in noiseless case.
We show that the condition above in fact implies $\rho > 0$ for the SC condition in \myref{cond2}, which is sufficient for accurate recovery even in the noisy case. In particular, with $\delta := \max_i \delta_i$, we have the following result:
\begin{theorem}[Structural Coherence (SC) Condition]
\label{prop:rho}
Let $\delta := \max_i \delta_i$ with $\delta_i$ as defined in \myref{eq:deltai}. If $\delta<1$, there exists a $\rho>0$ such that for any $\Delta_i \in \cC_i, i = 1, \ldots,k$, the SC condition in \myref{cond2} holds, i.e.,
\beq
\left\| \sum_{i=1}^k \Delta_i \right\|_2 \geq \rho \sum_{i=1}^k \| \Delta_i \|_2~.
\eeq
\end{theorem}

%
%

%
%

Thus, the SC condition is satisfied in the general case as long as the reflection $-\cC_i$ of any cone $\cC_i$ does not intersect, i.e., share a ray, with the Minkowski sum $\sum_{j \neq i} \cC_j$ of the other cones.






\section{Restricted Eigenvalue Condition for Superposition Models}
\label{sec:re}

Assuming that the SC condition is satisfied by the error cones $\{\cC_i\},i=1,\ldots,k$, in this section we show that the general RE condition in \myref{eqre} will be satisfied with high probability when the number of samples $n$ in the sub-Gaussian design matrix $X \in \mathbb{R}^{n \times p}$ crosses the sample complexity $n_0$. We give a precise characterization of the sample complexity $n_0$ in terms of the Gaussian width of the set $\cH$.


Our analysis is based on the results and techniques in \cite{trop15,mend14}, and we note that \cite{bcfs14} has related results using mildly different techniques. We start with a restricted eigenvalue condition on $\cH$. For a random vector $Z\in \mathbb{R}^p$, we define marginal tail function for an arbitrary set $E$ as
\beq
 Q_\xi (E;Z) = \inf_{u\in E}P(|\langle Z,u\rangle|\geq \xi)~,
\label{eq:mtf}
\eeq
noting that it is deterministic given the set $E\subseteq \mathbb{R}^p$.
%
%
Let $\epsilon_i,i=1,\ldots,n,$ be independent Rademacher random variables, i.e., random variable with  probability $\frac{1}{2}$ of being either $+1$ or $-1$, and let $X_i, i=1,\ldots,n,$ be independent copies of $Z$. We define empirical width of $E$ as
\beq
W_n(E;Z) = \sup_{u\in E}\langle h, u\rangle, \quad \text{where} \quad h = \frac{1}{\sqrt{n}} \sum_{i=1}^n \epsilon_i X_i~.
\label{eq:empw}
\eeq
With this notation, we recall the following result from \cite[Proposition~5.1]{trop15}:
\begin{lemma}
\label{lemsb}
Let $X\in \R^{n \times p}$ be a random design matrix with each row the independent copy of sub-Gaussian random vector $Z$. Then for any $\xi,\rho,t >0$, we have
\beq
\inf_{u\in\cH}\|X u\|_2\geq \rho\xi\sqrt{n} Q_{2\rho\xi}(\cH; Z) -2W_n(\cH;Z)-\rho\xi t
\eeq
with probability at least $1-e^{-\frac{t^2}{2}}$.
\end{lemma}
From Lemma \ref{lemsb}, in order to obtain lower bound of $\kappa$ in RE condition \myref{eqre}, we need to lower bound $Q_{2\rho\xi}(\cH; Z)$ and upper bound $W_n(\cH;Z)$. To lower bound $Q_{2\rho\xi}(\cH; Z)$, we consider the spherical cap
\begin{equation}
\textstyle \cA = (\sum_{i=1}^k\cC_i)\cap \cS^{p-1}~.
\label{eq:setA}
\end{equation}
From \cite{trop15,mend14}, one can obtain a lower bound to $Q_{\xi}(\cA;Z)$ based on the Paley-Zygmund inequality. The Paley-Zygmund inequality lower bound the tail distribution of a random variable by its second momentum. Let $u$ be an arbitrary vector, we use the following version of the inequality.
\beq\label{eq:pzi}
	P(|\langle Z,u\rangle|\geq 2\xi) \geq \frac{[\mathbb{E}|\langle Z,u\rangle|-2\xi]_+^2}{\mathbb{E} |\langle Z,u\rangle|^2}
\eeq
In the current context, the following result is a direct consequence of SC condition, which shows that $Q_{2\rho\xi}(\cH; Z)$ is lower bounded by $Q_{\xi}(\cA;Z)$, which in turn is strictly bounded away from 0.
The proof of Lemma~\ref{lemcp} is given in Appendix \ref{pf:lem:cp}.

\begin{lemma}
	\label{lemcp}
	Let sets $\cH$ and $\cA$ be as defined in \myref{eq:setC} and \myref{eq:setA} respectively. If the SC condition in \myref{cond2} holds, then the marginal tail functions of the two sets have the following relationship:
	\beq
	Q_{\rho\xi}(\cH;Z) \geq Q_{\xi} (\cA ; Z).
	\eeq
\end{lemma}


%

Next we discuss how to upper bound the empirical width $W_n(\cH; Z)$. Let set $\cE$ be arbitrary, and random vector $g\sim \cN(0, I_p)$ be a standard Gaussian random vector in $\mathbb{R}^p$. The Gaussian width \cite{bcfs14} of $\cE$ is defined as
\beq\label{eq:gwidth}
	w(E) = \mathbb{E}\sup_{u\in \cE}\langle g, u \rangle.
\eeq
Empirical width $W_n(\cH; Z)$ can be seen as the supremum of a stochastic process. One way to upper bound the supremum of a stochastic process is by generic chaining \cite{tala14,bcfs14,trop15}, and by using generic chaining we can upper bound the stochastic process by a Gaussian process, which is the Gaussian width.

As we can bound $Q_{2\rho\xi}(\cH; Z)$ and $W_n(\cH;Z)$, we come to the conclusion on RE condition. Let $X \in \R^{n \times p}$ be a random matrix where each row is an independent copy of the sub-Gaussian random vector $Z \in \mathbb{R}^p$, and where $Z$ has sub-Gaussian norm $\vertiii{Z}_{\psi_2} \leq \sigma_x$ \cite{vers12}. Let $\alpha = \inf_{u\in\cS^{p-1}} \mathbb{E}[|\langle Z,u\rangle|]$ so that $\alpha >0$ \cite{mend14,trop15}.
We have the following lower bound of the RE condition. The proof of Theorem \ref{thm:re} is based on the proof of \cite[Theorem~6.3]{trop15}, and we give it in appendix \ref{pf:re}.
\begin{theorem}[Restricted Eigenvalue Condition]
  Let $X$ be the sub-Gaussian design matrix that satisfies the assumptions above. If the SC condition \myref{cond2} holds with a $\rho > 0$, then with probability at least $1-\exp(-t^2/2)$, we have
  \beq
  \inf_{u \in \cH} \| X  u\|_2 \geq  c_1 \rho\sqrt{n} - c_2 w(\cH) - c_3 \rho t
  \eeq
  where $c_1, c_2$ and $c_3$ are positive constants determined by $\sigma_x$, $\sigma_\omega$ and $\alpha$.
  \label{thm:re}
\end{theorem}

To get a $\kappa>0$ in \myref{eqre},  one can simply choose $t = (c_1 \rho \sqrt{n} - c_2 w(\cH))/2c_3 \rho$. Then as long as $n > c_4 w^2(\cH)/\rho^2$ for $c_4 = c_2^2/c_1^2$, we have
\[
 \kappa = \inf_{u\in \cH}\frac{1}{\sqrt{n}}\|X u\|_2 \geq \frac{1}{2}\left(c_1 \rho - c_2 \frac{w(\cH)}{\sqrt{n}}\right) > 0,
\]
with high probability.

From the discussion above, if SC condition holds and the sample size $n$ is large enough, then we can find a matrix $X$ such that RE condition holds. On the other hand, once there is a matrix $X$ such that RE condition holds, then we can show that SC must also be true. Its proof is give in Appendix \ref{pf:prop:nec}.
\begin{proposition}\label{prop:nec}
	If $X$ is a matrix such that the RE condition \eqref{eqre} holds for $\Delta_i \in \cC_i$, then the SC condition \myref{cond2} holds.
\end{proposition}

Proposition~\ref{prop:nec} demonstrates that SC condition is a necessary condition for the possibility of RE. If SC condition does not hold, then there is $\{\Delta_i\}$ such that $\Delta_i \neq 0$ for some $i=1,\ldots,k$, but $\|\sum_{i=1}^k\Delta_i\|_2=0$ which implies $\sum_{i=1}^k\Delta_i = 0$. Then for every matrix $X$, we have $X\sum_{i=1}^k\Delta_i = 0$, and RE condition is not possible.

\section{General Error Bound}
\label{sec:desnoise}
Recall that the error bound in Theorem \ref{thdet} is given in terms of the noise-design (ND) interaction
\beq
s_n(\gamma) = \inf_{s>0}\left\{s:\sup_{u \in s\cC} \frac{1}{\sqrt{n}}\omega^T  X u \leq \gamma s^2 \sqrt{n}\right\}~.
\eeq
In this section, we give a characterization of the ND interaction, which yields the final bound on the componentwise error as long as $n \geq n_0$, i.e., the sample complexity is satisfied. 

Let $\omega$ be a centered sub-Gaussian random vector, and its sub-Gaussian norm $\vertiii{\omega}_{\psi_2} \leq \sigma_\omega$. Let $X$ be a row-wise i.i.d. sub-Gaussian random matrix, for each row $Z$, its sub-Gaussian norm $\vertiii{Z}_{\psi_2} \leq \sigma_x$. The ND interaction can be bounded by the following conclusion, and the proof of lemma \ref{lem:nd} is given in Appendix \ref{pf:lem:nd}.
\begin{lemma}\label{lem:nd}
Let design $X \in \R^{n \times p}$ be a row-wise i.i.d.~sub-Gaussian random matrix, and noise $\omega \in \R^n$
be a centered sub-Gaussian random vector. 
Then
$
	s_n(\gamma) \leq c\frac{ w(\overline{\cH})}{\gamma \sqrt{n}}~.
$
for some constant $c>0$ with probability at least
$1 - c_1\exp(-c_2 w^2(\overline{\cH})) - c_3\exp(-c_4n)$. Constant $c$ depends on $\sigma_x$ and $\sigma_\omega$.
\end{lemma}

In lemma \ref{lem:nd} and theorem \ref{eqre}, we need the Gaussian width of $\overline{\cH}$ and $\cH$ respectively. From definition, both $\overline{\cH}$ and $\cH$ is related to the union of different cones; therefore bounding the width of $\overline{\cH}$ and $\cH$ may be difficult. We have the following bound of $w(\cH)$ and $w(\bar{\cH})$ in terms of the width of the component spherical caps. The proof of Lemma \ref{lem:gw} is given in Appendix \ref{pf:lem:gw}.
\begin{lemma}[Gaussian width bound]\label{lem:gw}
	Let $\cH$ and $\overline{\cH}$ be as defined in \myref{eq:setC} and \myref{eq:setBC} respectively. Then, we have
	$
	w (\cH) = O \left( \max_i w (\cC_i \cap \cS_{p-1}) + \sqrt{\log k} \right)
	$
	and
	$
	w(\overline{\cH}) = O \left( \max_i w (\cC_i \cap B_{p}) + \sqrt{\log k} \right)
	$.
\end{lemma}

By applying lemma \ref{lem:gw}, we can derive the error bound using the Gaussian width of individual error cone. From our conclusion on deterministic bound in theorem \ref{thdet}, we can choose an appropriate $\gamma$ such that $\kappa^2 > \gamma$. Then, by combining the result of theorem \ref{thdet}, theorem \ref{thm:re}, lemma \ref{lem:nd} and lemma \ref{lem:gw}, we have the final form of the bound, as originally discussed in \myref{eq:bnd0}:
\begin{theorem}\label{th:bnd1}
For estimator \myref{eq:bnd0}, let $\cC_i=\cone \{ \Delta: R_i(\theta_i^*+\Delta) \leq R_i(\theta_i^*) \}$, design $X$ be a random matrix with each row an independent copy of sub-Gaussian random vector $Z$, noise $\omega$ be a centered sub-Gaussian random vector, and $B_p \subseteq\mathbb{R}^p$ be the centered unit euclidean ball. Suppose SC condition holds with
\[
\left\| \sum_{i=1}^k \Delta_i \right\|_2 \geq \rho \sum_{i=1}^k \| \Delta_i \|_2~.
\]
for any $\Delta_i \in \cC_i$ ans a constant $\rho >0$.
If sample size $n > c (\max_i w^2(\cC_i\cap S_{p-1}) + \log k )/\rho^2$, then with high probability,
\beq
\sum_{i=1}^k \| \hat{\theta_i}-\theta_i^* \|_2 \leq C \frac{\max_i w (\cC_i \cap B_{p}) + \sqrt{\log k}}{\rho^2\sqrt{n}},
\eeq
for constants $c, C>0$ that depend on sub-Gaussian norms $\vertiii{Z}_{\phi_2}$ and $\vertiii{\omega}_{\phi_2}$.
\end{theorem}

Thus, assuming the SC condition in \myref{cond2} is satisfied, the sample complexity and error bound of the estimator depends on the largest Gaussian width, rather than the sum of Gaussian widths. The result can be viewed as a direct generalization of existing results for $k=1$, when the SC condition is always satisfied, and the sample complexity and error is given by $w^2(\cC_1 \cap S_{p-1})$ and $w(\cC_1 \cap B_p)$ \cite{bcfs14, crpw12}.

\section{Accelerated Proximal Algorithm}
\label{sec:opt}
\begin{algorithm}[t!]
	\caption{Accelerated Proximal Algorithm}
	\begin{algorithmic}
		\STATE \textbf{Inputs:} $X$, $\mathbf{y}$.
		\STATE \textbf{Initialize:} $\{\theta_i^0\}_{i=1}^k = \{0\}$, $\alpha_0 = 0$, $\eta>0$, $0<\beta<1$.
		\FOR{$t = 0,1,\ldots,T$}
		\STATE Set $\eta^{t+1} = \eta$.
		\WHILE{true}
		\FOR { $i = 1, \ldots, k$ }
		\STATE $\tilde{\theta}_i^{t+1} = \Pi_{\Omega_i} ( \theta_i^{t} - \eta^{t+1} \nabla f_{\theta_i} ( \theta^t))$
		\ENDFOR
		\IF{ $f ( \tilde{\theta}^{t+1}) \leq f (  \theta^t
			) + \nabla^T f ( \theta^t ) ( \tilde{\theta}^{t+1} -  \theta^t) + \frac{1}{2
				\eta^{t+1}} ( \sum_{i=1}^k \| \tilde{\theta}_i^{t+1} - \theta_i^t \|_2^2 )$}
		\STATE break
		\ENDIF
		\STATE $\eta^{t+1} = \beta \eta^{t+1}$
		\ENDWHILE
		\STATE $\alpha_{t + 1} = \frac{1 + \sqrt{1 + 4 \alpha_t^{2}}}{2}, \theta_i^{t+1} = \tilde{\theta}_i^{t + 1} +
		\frac{\alpha_t - 1}{\alpha_{t + 1}} ( \tilde{\theta}_i^{t + 1} - \theta_i^t)$
		\ENDFOR
	\end{algorithmic}\label{algdm}
\end{algorithm}

In this section, we propose a general purpose algorithm for solving problem \eqref{eq:est1}.
For convenience, with $\theta = \sum_{i=1}^k\theta_i$, we set $f(\theta) = f(\sum_{i=1}^k \theta_i) = \| \mathbf{y} - X \theta \|_2^2$ and $\Omega_i = \{ \theta_i | R_i(\theta_i ) \leq R_i(\theta_i^*) \}$. While the norms $R_i(.)$ may be non-smooth, one can design a general algorithm as long as the proximal operators $\Pi_{\Omega_i}(v) = \argmin_{u\in\Omega_i} \|u - v\|_2^2$ for each set $\Omega_i$ can be efficiently computed. The algorithm is simply the proximal gradient method
\cite{pabo14}, where each component $\theta_i$ is cyclically updated in each iteration (see Algorithm~\ref{algdm}):
\begin{equation}
\begin{split}
\tilde{\theta}_i^{t+1} & = \underset{\theta_i\in\Omega_i}{\argmin}~~ \lip\nabla_{\theta_i} f(\theta^t), \theta_i - \theta_i^t  \rip+ \frac{1}{2\eta^{t+1}}\|\theta_i - \theta_i^t\|_2^2~,\\
 & = \underset{\theta_i\in\Omega_i}{\argmin} ~~\|\theta_i - (\theta_i^t - \eta^{t+1} \nabla_{\theta_i} f(\theta^t))\|_2^2~,
\end{split}
\end{equation}
where $\eta^{t+1}$ is the learning rate. To determine a proper $\eta^{t+1}$, we use a  backtracking step \cite{bete09}. Starting from a constant $\eta^{t+1} = \eta$, in each step we first update $\tilde{\theta}_i^{t+1}$; then we decide whether $\tilde{\theta}_i^{t+1}$ satisfies condition:
\beq\label{eq:armijo}
f ( \tilde{\theta}^{t+1}) \leq f (  \theta^t
) + \nabla^T f ( \theta^t ) ( \tilde{\theta}^{t+1} -  \theta^t) + \frac{1}{2
	\eta^{t+1}} ( \sum_{i=1}^k \| \tilde{\theta}_i^{t+1} - \theta_i^t \|_2^2 ).
\eeq
If the condition \eqref{eq:armijo} does not hold, then we decrease $\eta^{t+1}$ till \eqref{eq:armijo} is satisfied.
%
%
%
Based on existing results \cite{bete09}, the basic method can be accelerated by setting the starting point of the next iteration $\theta_i^{t+1}$ as a proper combination of $\tilde{\theta}_i^{t+1}$ and $\theta_i^t$. By \cite{bete09}, one can use the updates:
\beq
\theta_i^{t+1} = \tilde{\theta}_i^{t + 1} + \frac{\alpha_t - 1}{\alpha_{t + 1}} ( \tilde{\theta}_i^{t + 1} - \theta_i^t)~, \quad \text{where} \quad \alpha_{t + 1} = \frac{1 + \sqrt{1 + 4 \alpha_t^2}}{2}~.
\eeq

Convergence of Algorithm \ref{algdm} has been studied in \cite{bete09}. The backtracking step ensures that the convergence of algorithm \ref{algdm}. The work \cite{bete09} also give the convergence rate of Algorithm \ref{algdm}, which is $O(1/t^2)$. Therefore, we can always reach a stationary point of problem \eqref{eq:est1} using Algorithm \ref{algdm}.

\section{Noiseless Case: Comparing Estimators}
\label{sec:related}
In this section, we present a comparative analysis of estimator 
\beq\label{eq:est2}
 \min_{\{ \theta_i \}} \sum_{i=1}^k \lambda_iR_i(\theta_i) \quad s.t. \quad X\sum_{i=1}^{k} \theta_i = y
\eeq
with the proposed estimator \myref{eq:est1} in the noiseless case, i.e., $\omega = 0$. In essence, we show that the two estimators have similar recovery conditions, but the existing estimator \myref{eq:est2} needs additional structure for unique decomposition of $\theta$ into the components $\{ \hat{\theta}_i\}$.




\begin{figure}[t]
	\centering
	\includegraphics[width=0.6\textwidth]{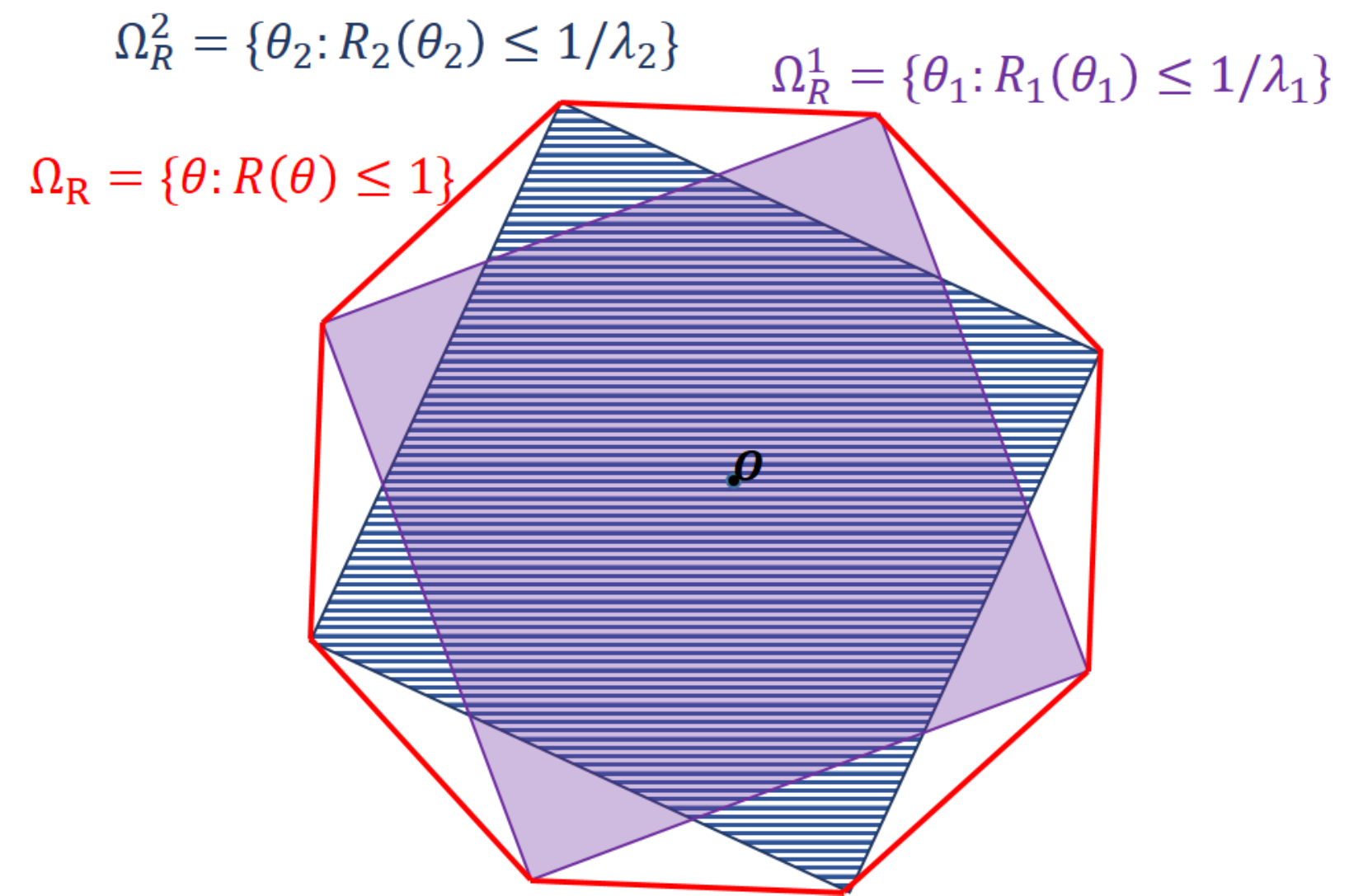}
	\caption{The relationship of different norm balls when $k=2$. The blue and purple polygons are the norm ball of norms $R_1(.)$ and $R_2(.)$ respectively. The red line is the outline of $R(.)$ norm ball. Note that for any point in the red line, we will be able to decompose it to the two vertexes around it.}
\end{figure}

The estimator \myref{eq:est2} needs to consider the so-called ``infimal convolution'' \cite{rock70,yara12} over different norms to get a (unique) decomposition of $\theta$ in terms of the components $\{\hat{\theta}_i\}$. Denote
\begin{equation}\label{defic}
  R(\theta)=\min_{\{ \theta_i \}:\sum_i\theta_i=\theta} \sum_{i=1}^k \lambda_iR_i(\theta_i)~.
\end{equation}
Results in \cite{rock70} show that (\ref{defic}) is also a norm. Thus estimator \myref{eq:est2} can be rewritten as
\beq\label{defic2}
  \min_{\theta} R(\theta)~s.t.~X\theta = y.
\eeq
Interestingly, the above discussion separates the estimation problem in $\myref{eq:est2}$ into two parts---solving \myref{defic2} to get $\hat{\theta}$, and then solving \myref{defic} to get the components $\{ \hat{\theta}_i \}$. The problem (\ref{defic2}) is a simple structured recovery problem, and is well studied \cite{crpw12,trop15}. Using infimal convolution based decomposition problem \myref{defic}
to get the components $\{ \hat{\theta}_i\}$ will be our focus in the sequel.






To get some properties of decomposition \eqref{defic}, we consider the unit norm balls for norm $R(.)$ and component norms $R_i(.)$:
\[
 \Omega_R = \{\theta \in \R^p: R(\theta)\leq 1\}\quad \text{and}\quad
 \Omega_R^i = \{\theta_i \in \R^p: R_i(\theta_i) \leq 1\}~, i=1,\ldots,k~.
\]
The norm balls are related by the following result, we give the proof in appendix \ref{pf:lemic}.
\begin{lemma}
For a given set $\{\lambda_i\}$, the infimal convolution norm ball $\Omega_R$ is the convex hull of
$\bigcup_{i=1}^k \frac{1}{\lambda_i}\Omega_R^i$, i.e., $\Omega_R = \conv(\bigcup_{i=1}^k \frac{1}{\lambda_i}\Omega_R^i)$.
\label{lemic}
\end{lemma}


%


\begin{figure}{t}
	\centering
	\subfigure[]{
		\includegraphics[width=0.45\textwidth]{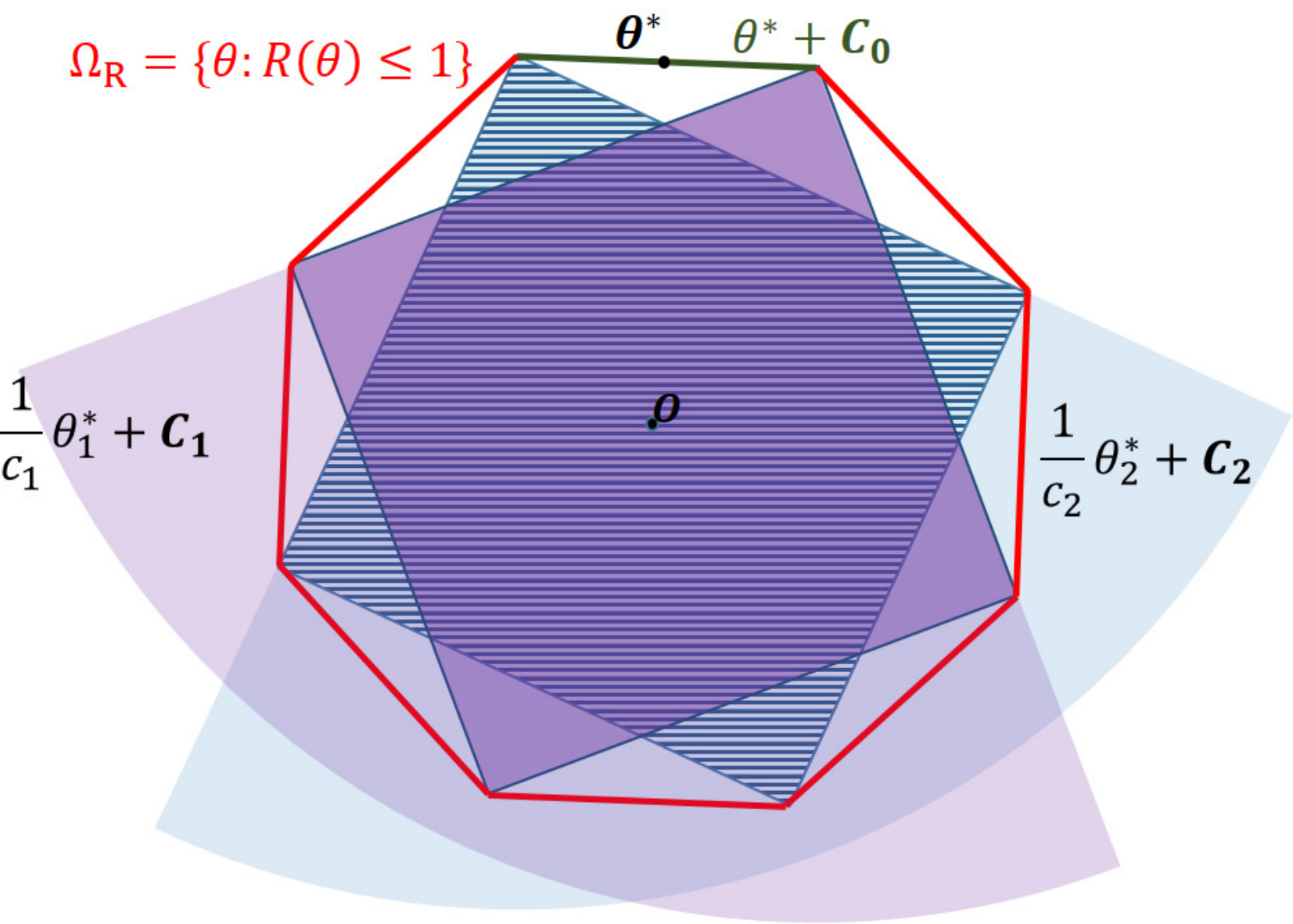}
	}
	\subfigure[]{
		\includegraphics[width=0.45\textwidth]{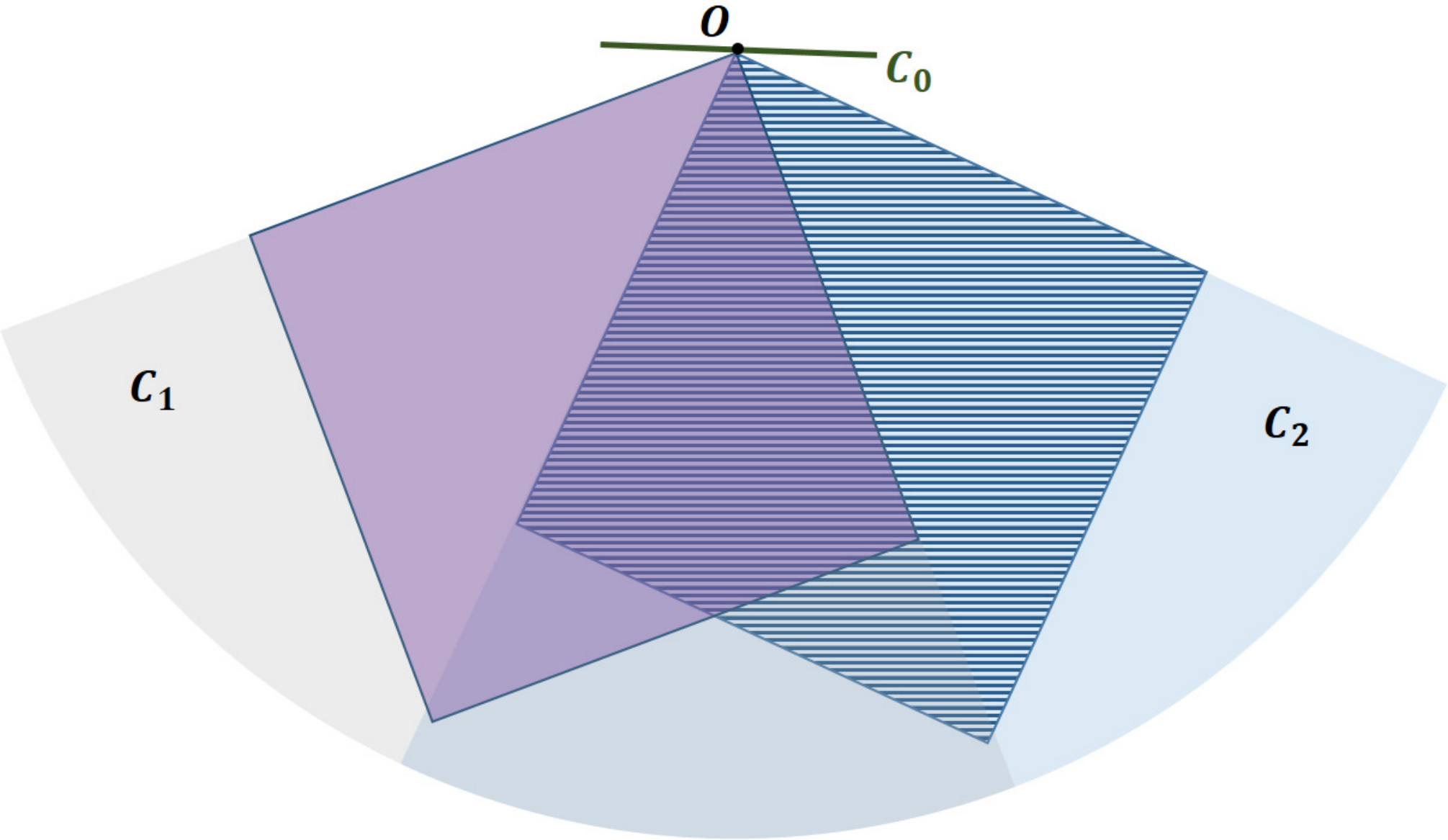}
	}
	\caption{Consider the case when $k=2$. Let $c_i = \lambda_iR_i(\theta_i)$ for $i=1,2$.
	\textbf{(a)} is the structure of error around the true value $\theta^*$. The green segment $\cC_0$ is a subspace determined by $\theta_1^*$ and $\theta_2^*$. For the superposition in (a), error of $\theta^*$ is composed of three parts: $\theta^*+\cC_0$, $\frac{1}{c_1}\theta_1^* + \cC_1$ and $\frac{1}{c_2}\theta_2^* + \cC_2$. In \textbf{(b)}, we move the green segment and two error cones to the origin, then the uniquely recovery condition is that if we reflect one of the three structures, their intersection remains $\{0\}$.}
\end{figure}

Lemma \ref{lemic} illustrates what the decomposition (\ref{defic}) should be like. If $\theta$ is a point
on the surface of the norm ball $\Omega_R$, then the value of $R(\theta)$ is the convex combination
of some $\theta_i$ on the surface of $\frac{1}{\lambda_i}\Omega_R^i$ such that $R_i(\theta_i) = R(\theta)$.  
Hence if $\theta$ can be successfully decomposed into different components along the direction of $\theta_i$, then we should be able to connect $\theta_i$ and $\theta$ by a surface on the $R(.)$ norm ball, or they have to be ``close''.  
Interestingly, the above intuition of ``closeness'' between different components $\theta_i$ can be described in the language of cones, in a way similar to the structural coherence property discussed in Section~\ref{sec:geometry}.

Given the intuition above, we state the main result in this section below. Its proof is given in appendix \ref{pf:thpa}.
\begin{theorem}\label{thpa}
	Given $\hat{\theta}_1, \ldots, \hat{\theta}_k$ and define
	\beq\label{def:c0}
	\cC_0 = \left\{\sum_{\theta_i\neq 0} \left(\frac{c_i'}{c_i}-1\right)\theta_i~|~c_i'\geq 0, \sum_{i=1}^k c_i'=1\right\}.
	\eeq
	Suppose $dim(span\{\theta_i\})=k$,
	 then there exist $\lambda_1, \ldots, \lambda_k$ such that $\sum_{i=1}^k \hat{\theta}_i = \theta$ are unique solutions of (\ref{defic}) if and only if there are $c_1,\ldots,c_k$ with $c_i\geq 0$ and $\sum_{i=1}^kc_i = 1$ such that for the corresponding error cone $\cC_i$ of $\hat{\theta}_i$ and $C_0$ defined above,
	$
- \cC_i\cap \sum_{j \neq i} \cC_j = \{ 0 \},
	$
for $i=0,1,\ldots,k$.
\end{theorem}


%
%


Theorem~\ref{thpa} illustrate that the successful decomposition of ($\ref{defic}$)
requires an additional condition, i.e., $-\cC_0\cap \sum_{i=1}^k \cC_i = \{0\}$ beyond that is needed by the SC condition (see Section~\ref{sec:geometry}).
The additional condition needs us to choose parameters $\{\lambda_i\}$ properly. Theorem \ref{thpa} shows that $\{\lambda_i\}$ depends on both $\{\theta_i^*\}$ and $\{c_i\}$. For appropriate $\{\theta_i^*\}$, there may be a range of $\{c_i\}$ such that the solution is unique.
Therefore, in noiseless situation, if we know $\{R_i(\theta_i^*)\}$, then solving estimator \myref{eq:est2} would be a better idea, because it requires less condition to recover the true value and we do not need to choose parameters $\{\lambda_i\}$. 


%

\section{Related Work}
\label{sec:related2}
Structured superposition models have been studied in recent literatures. Early work focus on the case when k=2 and noise $\omega=0$, and assume specific structures such as sparse+sparse \cite{dohu01}, and low-rank+sparse \cite{cspw11}. \cite{hskz11} analyze error bound for low-rank and sparse matrix decomposition with noise. Recent work have considered more generalized models and structures. \cite{agnw12} analyze the decomposition of a low-rank matrix plus another matrix with generalized structure. \cite{foma14} propose an estimator for the decomposition of two generalize structured matrices, while one of them has a random rotation. Because of the increase in practical application and non-trivial of such problem, people have begun to work on unified frameworks for superposition model. In \cite{wgmm12}, the authors generalize the noiseless matrix decomposition problem to arbitrary number of superposition under random orthogonal measurement. \cite{yara12} consider the superposition of structures of structures captured by decomposable norm, while \cite{mctr13} consider general norms but with a different measurement model, involving componentwise random rotations. These two papers are similar in spirit to our work, so we briefly discuss and differentiate our work from these papers.

\cite{yara12} consider a general framework for superposition model, and give a high-probability bound for the following estimation problem:
\beq\label{dme}
\min_{\theta_i,i=1,\ldots,k}\left\|y - X \sum_{i=1}^{k}\theta_i\right\|_2^2 + \sum_{i=1}^{k}\lambda_i R_i(\theta_i)
\eeq
they assume each $R_i()$ to be a special kind of norm called decomposable norm. the authors used a different approach for RE condition. They decompose $\|X \sum_{i=1}^k \Delta_i\|_2$ into two parts. One is
\beq\textstyle
\frac{1}{n}\|X\Delta_i\|_2^2 \geq \kappa\|\Delta_i\|_2^2,
\eeq
which characterizes the restricted eigenvalue of each error cone. The other is
\beq\label{cond:int}\textstyle
\frac{2}{n}|\sum_{i<j}\langle X\Delta_i, X\Delta_j\rangle| \leq \frac{\kappa}{2} \sum_{i=1}^{k} \|\Delta_i\|_2^2,
\eeq
which characterizes the interaction between different error cones. \myref{cond:int} is a strong assumption, and RE condition can hold without it. If $\Delta_i$ and $\Delta_j$ are positively correlated, then large interaction terms will make our RE condition stronger. Therefore their results are restricted.

\cite{mcco13} consider an estimator like \eqref{eq:est1}, which is
\beq\label{rre}
\min_{\theta_i,i=1,\ldots,k}\left\|y - X \sum_{i=1}^{k}Q_i\theta_i\right\|_2^2 ~ \text{s.t.}~ R_i(\theta_i) \leq R_i (\theta_i^{\ast}),~i = 1, \ldots, k,
\eeq
where $Q_i$ are known random rotations. Problem \myref{rre} is then transformed into a geometric problem: whether $k$ random cones intersect. The componentwise random rotation can ensure that any kind of combination can be recovered with high probability. However, in practical problems, we need not have such random rotations available as part of the measurements. Further, their analysis is primarily focused on the noiseless case.




\section{Application of General Bound}
\label{sec:examples}
In this section, we instantiate the general error bounds on Morphological Component Analysis (MCA), and low-rank and sparse matrix decomposition. The proofs are provided in appendix \ref{pf:ex}.


\subsection{Morphological Component Analysis Using $l_1$ Norm}
In Morphological Component Analysis \cite{dohu01}, we consider the following linear model 
\[
	y = X(\theta_1^* + \theta_2^*) + \omega,
\]
where vector $\theta_1$ is sparse and vector $\theta_2$ is sparse under a rotation $Q$.
In \cite{dohu01}, the authors introduced a quantity 
\beq\label{eq:ic}
M = \max_{i,j}|Q_{ij}|.
\eeq
For small enough $M$, if the sum of their sparsity is lower than a constant related to $M$, we can recovery them. We show that for two given sparse vectors, our SC condition is more general.

Consider the following estimator
\beq
	\min_{\theta_1,\theta_2}\|y - X(\theta_1 + \theta_2)\|_2^2 \quad
	s.t. \quad \|\theta_1\|_1\leq \|\theta_1^*\|_1, \|Q\theta_2\|_1 \leq \|Q\theta_2^*\|_1,
\label{lle}
\eeq
where vector $y\in \mathbb{R}^n$ is the observation, vectors $\theta_1, \theta_2 \in \mathbb{R}^p$ are the parameters we want to estimate, matrix $X\in \mathbb{R}^{n\times p}$ is a sub-Gaussian random design, matrix $Q\in \R^{p\times p}$ is orthogonal. We assume $\theta_1$ and $Q\theta_2$ are $s_1$-sparse and $s_2$-sparse vectors respectively. Function $\|Q.\|_1$ is still a norm.

Suppose $s_1 = 1$, $s_2 = 1$, and the i-th entry of $\theta_1$ and the j-th entry of $Q\theta_2$ are non-zero. If 
\[
Q_{ij}\sign(\theta_1^*)_i\sign(Q\theta_2^*)>0,
\]
then we have
\beq\label{eq:bndr}
 \rho \geq \sqrt{(1 - \sqrt{1 - Q_{ij}^2})/2}.
\eeq
Thus we will have chance to separate $\theta_1$ and $\theta_2$ successfully.
It is easy to see that $M$ is lower bounded by $\theta_1^T Q \theta_2$. Large $\theta_1^T Q \theta_2$ leads to larger $M$, but also leads to larger $\rho$, which is better for separating $\theta_1$ and $\theta_2$. The proof of above bound of $\rho$ is given in Appendix \ref{pf:sc11}.

In general, it is difficult for us to derive a lower bound of $\rho$ like \ref{eq:bndr}. Instead, we can derive the following sufficient condition in terms of $M$:

\begin{theorem}\label{th:mcal1}
	If $M \leq  \frac{1}{8\sqrt{s_1s_2}}$, then for problem \eqref{lle} with high probability
	\[
	\|\theta_1 - \theta_1^*\|_2 + \|\theta_2 - \theta_2^*\|_2  = O\left( \max\left\{\sqrt{\frac{s_1\log p}{n}}, \sqrt{\frac{s_2\log p}{n}}\right\} \right).
	\]
\end{theorem}
When $s_1=s_2=1$, this condition $M \leq \frac{1}{8\sqrt{s_1s_2}}$ is much stronger than $\eqref{eq:bndr}$, because every entry of $Q$ has to be smaller than $1/8$;

\subsection{Morphological Component Analysis Using $k$-support Norm}

$k$-support norm \cite{arfs12} is another way to induce sparse solution instead of $l_1$ norm. Recent works \cite{arfs12,chcb14} have shown that $k$-support norm has better statistical guarantee than $l_1$ norm. For arbitrary $\theta \in \mathbb{R}^p$, its $k$-support norm  $\|\theta\|_k^{sp}$ is defined as
\[
	\|\theta\|_k^{sp} = \inf\left\{\sum_{I\in\mathcal{G}_k} \|u_I\|_2 : \supp(u_I)\subseteq I, \sum_{I\in\mathcal{G}_k}u_I = \theta\right\}.
\]

For the superposition of an $s_1$ sparse vector and an $s_2$ sparse vector, the best choice is to use $s_1$-support norm and $s_2$-support norm. The new problem is
\beq\label{kke}
	\min_{\theta_1,\theta_2}\|y - X(\theta_1 + \theta_2)\|_2^2 \quad
	s.t. \quad \|\theta_1\|_{s_1}^{sp}\leq \|\theta_1^*\|_{s_1}^{sp}, \|Q\theta_2\|_{s_2}^{sp} \leq \|Q\theta_2^*\|_{s_2}^{sp}.
\eeq
Denote $\sigma_{s_1,s_2}(Q)$ as the set of all the largest singular values of $Q$'s $s_1\times s_2$ submatrices. Let $\sigma = \max\sigma_{s_1,s_2}(Q)$. In this case, we have the following sufficient condition and high probability error bound:
\begin{theorem}\label{th:mcakk}
	If $\sigma \leq \frac{1}{4(1+\frac{\theta_{1max}}{\theta_{1min}})(1+\frac{\theta_{2max}}{\theta_{2min}})}$ where $\theta_{max} = \max_{i\in \supp(\theta)}|\theta_i|$ and $\theta_{min} = \min_{i\in \supp(\theta)}|\theta_i|$, then we have for problem \eqref{kke} with high probability
	\[
	\|\theta_1 - \theta_1^*\|_2 + \|\theta_2 - \theta_2^*\|_2 = O\left(\max\left\{\sqrt{\frac{s_1\log(p-s_1)}{n}},\sqrt{\frac{s_2\log(p-s_2)}{n}}\right\}\right).
	\]
\end{theorem}

In the problem setting of theorem \ref{th:mcakk}, both norms are not decomposable. Therefore we can not apply the framework of \cite{yara12} for this problem.


\subsection{Low-rank and Sparse Matrix Decomposition}
To recover a sparse matrix and low-rank matrix from their sum \cite{clmw11,cspw11}, one can use $k$-support norm \cite{arfs12} to induce sparsity and nuclear norm to induce low-rank.
These two kinds of norm ensure that the sparsity and the rank of the estimated matrices are small.
If we have $k > 1$, the framework in \cite{yara12} is not applicable, because $k$-support norm is not decomposable. When $k=1$, the component norms simplify to $\| \cdot \|_1$ for sparsity.

Suppose we have a rank-$r$ matrix $L^*$ and a sparse matrix $S^*$ with $s$ nonzero entries, $S^*,L^*\in R^{d_1\times d_2}$. Our observation $Y$ comes from  the following problem
\[
Y_i = \langle X_i, L^*+S^* \rangle + E_i, i=1, \ldots, n,
\]
where each $X_i \in \mathbb{R}^{d_1\times d_2}$ is a sub-Gaussian random design matrix. $E_i$ is the noise matrix. We want to recover $S^*$ and $L^*$ using $s$-support norm and nuclear norm respectively, so that the estimator takes the form:
\beq\label{lsd}
\min_{L,S} \sum_{i=1}^n (Y_i - \langle X_i, L + S \rangle)^2\quad
s.t. \quad \vertiii{ L}_*\leq \vertiii{ L^*}_*, \|S\|_s^{sp} \leq \|S^*\|_s^{sp}.
\eeq
By using Theorem~ \ref{th:bnd1}, and existing results on Gaussian widths, the error bound is given by

\begin{theorem}\label{th:rpca}
	If there is a $\rho>0$ for problem \eqref{lsd}, then with high probability
	\[
	\|L - L^*\|_2 + \|S - S^*\|_2 = O\left(\max\left\{\sqrt{\frac{s\log(d_1d_2-s)}{n}}, \sqrt{\frac{r(d_1+d_2-r)}{n}}\right\} \right).
	\]
\end{theorem}

Theorem \ref{th:rpca} requires SC condition to hold. When will SC condition for \myref{lsd} holds? Early work have shown that to successfully estimate both $L$ and $S$, the low-rank matrix $L$ should satisfy ``incoherence'' condition \cite{clmw11}. From example in Appendix \ref{ex:rpca11}, we can recovery matrix $L$ and $S$ even incoherence condition does not hold.


\section{Experimental Results}
\label{sec:expr}
In this section, we confirm the theoretical results in this paper with
some simple experiments. We show our experimental results under different settings.
In our experiments we focus on MCA when $k=2$. The design matrix $X$ are generated from Gaussian distribution such that every entry of $X$ subjects to $\mathcal{N}(0,1)$. The noise $\omega$ is generated from Gaussian distribution such that every entry of $\omega$ subjects to $\mathcal{N}(0,1)$. We implement our algorithm ~\ref{algdm} in MATLAB. We use synthetic data in all our experiments, and let the true signal
\[
\theta_1 = (\underbrace{1,\ldots,1}_{s_1},0\ldots,0),Q\theta_2 = (\underbrace{1,\ldots,1}_{s_2},0\ldots,0)
\]
We generate our data in different ways for our three experiments.

\subsection{Recovery From Noisy Observation}
\begin{figure}[t]
	\begin{center}
		\includegraphics[width=0.7\textwidth]{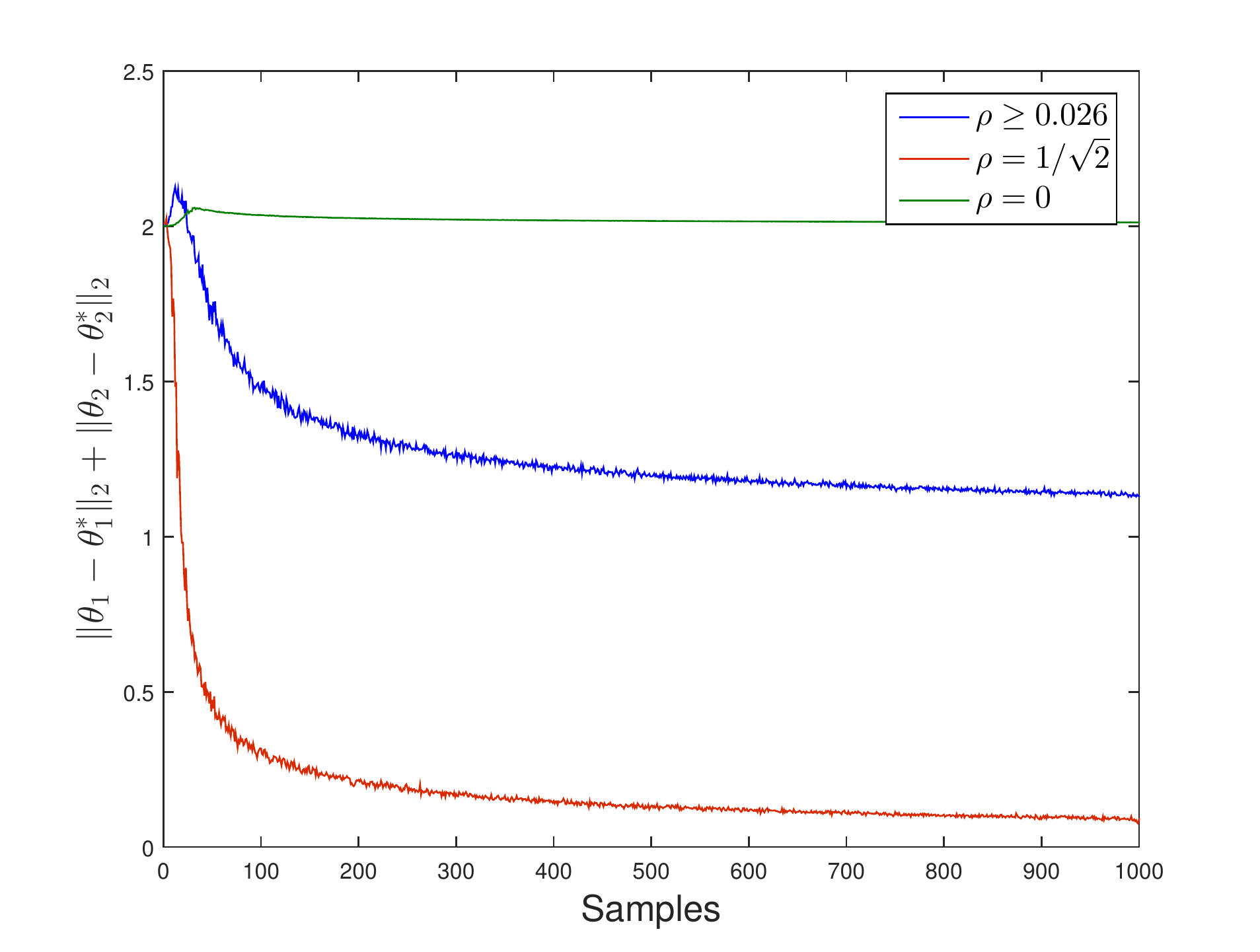}
		\caption{Effect of parameter $\rho$ on estimation error when noise $\omega\neq 0$. We choose the parameter $\rho$ to be $0$, $1/\sqrt{2}$, and a random sample.}\label{plt:mca:noise}
	\end{center}
\end{figure}

In our first experiment, we test the impact of $\rho$ on the estimation error. We choose three different matrices $Q$, and $\rho$ is determined by the choice of $Q$. The first $Q$ is given by random sampling: we sample a random orthogonal matrix $Q$ such that $Q_{ij}>0$, and $\rho$ is lower bounded by \myref{eq:bndr}. The second and third $Q$ is given by identity matrix $I$ and its negative $-I$; therefore $\rho=1/\sqrt{2}$ and $\rho=0$ respectively. We choose dimension $p=1000$, and let $s_1 = s_2 = 1$. The number of samples $n$ varied between 1 and 1000. Observation $y$ is given by $y = X(\theta_1^* + \theta_2^*) + \omega$. In this experiment, given $Q$, for each $n$, we generate 100 pairs of $X$ and $w$. For each $(X,w)$ pair, we get a solution $\hth_1$ and $\hth_2$. We take the average over all $\| \hth_1 - \theta_1^{\ast} \|_2 + \| \hth_2 - \theta_2^{\ast} \|_2$. Figure \ref{plt:mca:noise} shows the plot of number of samples vs the average error. From figure \ref{plt:mca:noise}, we can see that the error curve given by random $Q$ lies between curves given by two extreme cases, and larger $\rho$ gives lower curve.

\subsection{Recovery From Noiseless Observation}
\begin{figure}[tb]
	\vspace{-2mm}
	\begin{center}
		\includegraphics[width=0.7\textwidth]{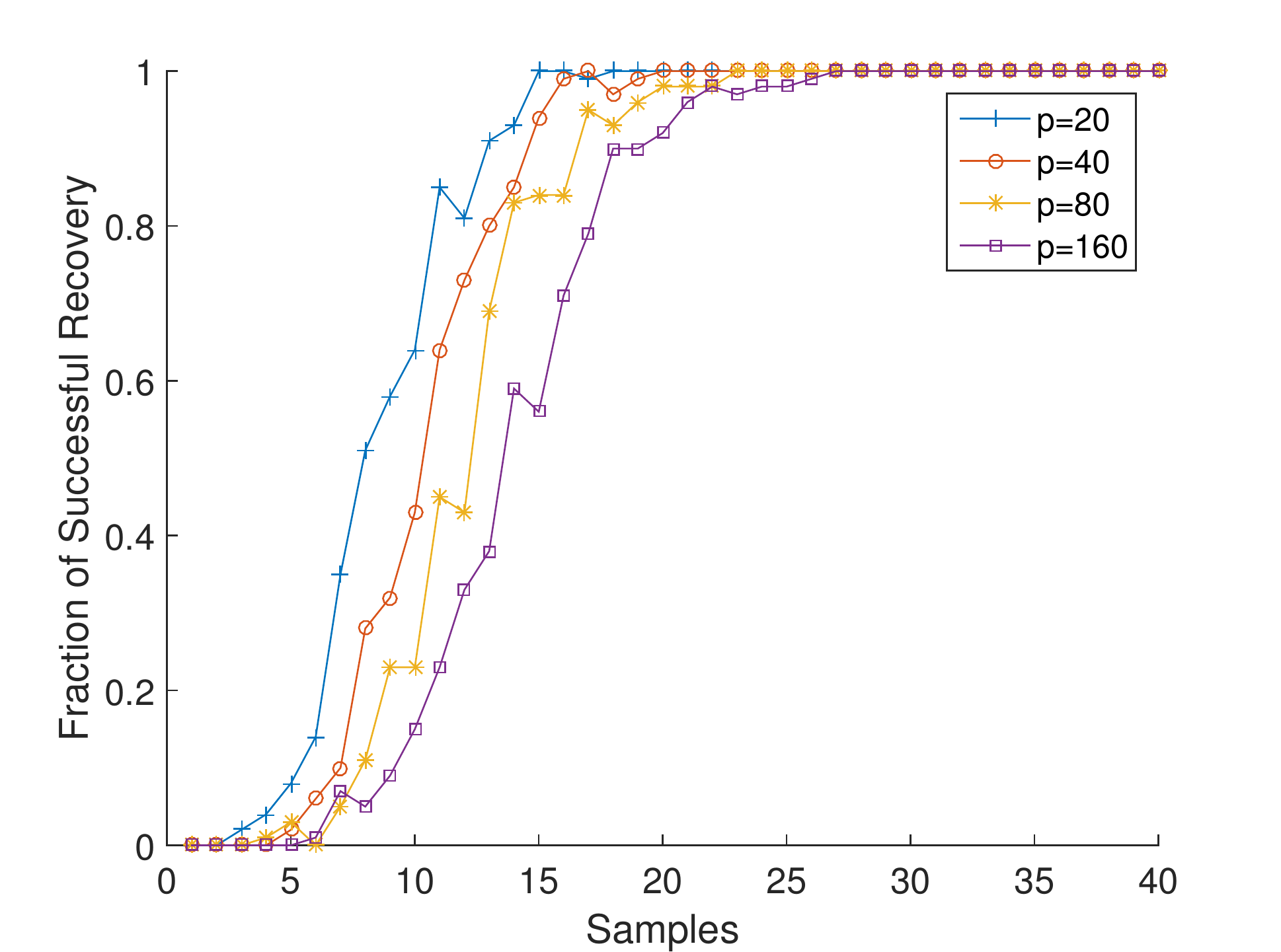}
		\caption{Effect of dimension $p$ on fraction of successful recovery in noiseless case. Dimension $p$ varies in $\{20, 40, 50, 150\}$}
		\label{plt:mca:rnd}
	\end{center}
	\vspace{-10mm}
\end{figure}

In our second experiment, we test how the dimension $p$ affects the successful recovery of true value. In this experiment, we choose different dimension $p$ with $p=20$, $p=40$, $p=80$, and $p=160$. We let $s_1 = s_2 = 1$. To avoid the impact of $\rho$, for each sample size $n$, we sample $100$ random orthogonal matrices $Q$. Observation $y$ is given by $y = X(\theta_1^*+\theta_2^*)$. For each solution $\hth_1$ and $\hth_2$ of \myref{lle}, we calculate the proportion of $Q$ such that $\| \hth_1 - \theta_1^{\ast} \|_2 + \| \hth_2 - \theta_2^{\ast} \|_2 \leq 10^{-4}$. We increase $n$ from $1$ to $40$, and the plot we get is figure \ref{plt:mca:rnd}. From figure \ref{plt:mca:rnd} we can find that the sample complexity required to recover $\theta_1^*$ and $\theta_2^*$ increases with dimension $p$.

\subsection{Recovery using $k$-support norm}

\begin{figure}[tb]
	\vspace{-2mm}
	\begin{center}
		\includegraphics[width=0.7\textwidth]{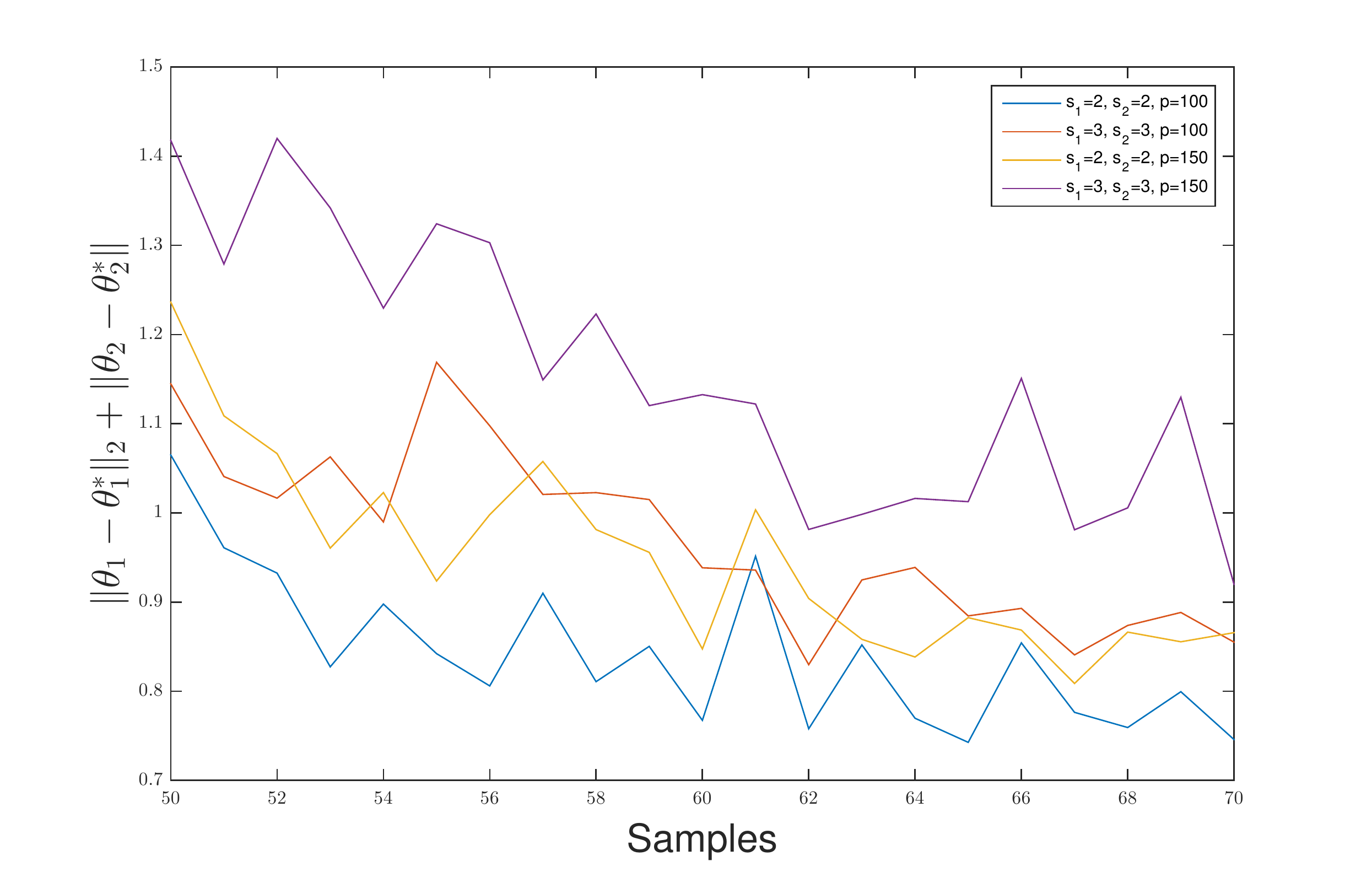}
		\caption{Effect of sparsity $s_1$, $s_2$ on estimation error. In all cases we use $k$-support norm instead of $l1$ norm.}
		\label{plt:mca:ksupp}
	\end{center}
	\vspace{-10mm}
\end{figure}

In our last experiment, we test the impact of sparsity on the estimation error. In this experiment we solve problem \eqref{kke}, and let both $s_1$ and $s_2$ vary from 2 to 3. We set the matrix $Q$ to be a $p\times p$ discrete cosine transformation (DCT) matrix \cite{dohu01}. We use different problem size with $p=100$ and $p=150$. Number of samples $n$ varies from $30$ to $70$. For each $n$, we generate 20 pairs of $X$ and $\omega$. For each $(X,w)$ pair, we get a solution $\hth_1$ and $\hth_2$. We take the average over all $\| \hth_1 - \theta_1^{\ast} \|_2 + \| \hth_2 - \theta_2^{\ast} \|_2$. The plot is shown in figure \ref{plt:mca:ksupp}. From figure \ref{plt:mca:ksupp}, we can see that the error curve increases as dimensionality and sparsity increases. If dimensionality $p$ is fixed, then lower sparsity implies better estimation result.

\section{Conclusions}
\label{sec:conc}

We present a simple estimator for general superposition models and give a purely geometric characterization, based on structural coherence, of when accurate estimation of each component is possible. Further, we establish sample complexity of the estimator and upper bounds on componentwise estimation error and show that both, interestingly, depend on the largest Gaussian width among the spherical caps induced by the error cones corresponding to the component norms. Going forward, it will be interesting to
investigate specific component structures which satisfy structural coherence, and also extend
our results to allow more general measurement models.



{\bf Acknowledgements:}\quad
The research was also supported by NSF grants IIS-1563950, IIS-1447566, IIS-1447574, IIS-1422557, CCF-1451986, CNS- 1314560, IIS-0953274, IIS-1029711, NASA grant NNX12AQ39A, and gifts from Adobe, IBM, and Yahoo.

\bibliographystyle{plain}
\bibliography{dirty_model}

\section*{Appendix}

\appendix


\section{Proof of Theorem \ref{thdet}}\label{pf:th:det}
\begin{theorem}[Deterministic bound]
	Assume that the RE condition in \myref{eqre} is satisfied in $\cC$ with parameter $\kappa$. Then, if $\kappa^2 > \gamma$, we have
	$
	\sum_{i=1}^k \|\Delta_i\|_2 \leqslant 2s_n(\gamma)
	$.
\end{theorem}
\proof
	By feasibility of $\theta^*$ and optimality of $\hat{\theta}$, we have
	\[
	\|Y - X\hat{\theta}\|_2^2 \leq \|Y - X\theta^*\|_2^2~.
	\]
	If $\hat{\Th} = \sum_i \hat{\theta}_i$ is an optimum of
	\eqref{eq:est1}, we have
	\[ 
    \| Y - X \hat{\theta} \|_2^2 = \| X (\hat{\theta} - \theta^*) \|_2^2	- 2 \omega^T X (\hat{\theta} - \theta^{\ast}) + \| \omega \|_2^2 .
    \]
	With $\Delta = \hat{\theta} - \theta^{\ast}$, $\Delta_i = \hat{\theta}_i -
	\theta_i^{\ast}$, we have
	\begin{equation}
	\| Y - X \hat{\theta} \|_2^2 - \| Y - X \theta^{\ast} \|_2^2 = \| X
	\Delta \|_2^2 - 2 \omega^T X \Delta \leqslant 0.
	\label{eq:opt1}
	\end{equation}
	For any $\Delta = \sum_{i=1}^k
	\Delta_i, \Delta_i \in \cC_i$, for the sake of contraction let $\sum_{i=1}^k \|\Delta_i\|_2\geq 2s_n(\gamma)$.
	Then we have
	\begin{equation}\textstyle
	\frac{1}{n} (\| X \Delta \|_2^2 - 2 \omega^T X \Delta) \geq \left(\sum_{i=1}^k
	\| \Delta_i \|_2 \right)^2 (\kappa^2 - \gamma) > 0.
	\end{equation}
	since $\kappa^2 > \gamma$. However, the inequality contradicts \myref{eq:opt1}.
	Therefore $\sum_{i=1}^k \| \Delta_i \|_2 \leq 2s_n(\gamma)$. \qed

\section{Geometry of Structural Coherence}
In this section, our goal is to characterize the geometric property of our SC condition. We start from a simple case when $k=2$.
	
\subsection{Proof of Lemma \ref{lemta}}
\begin{lemma}
	If there exists a $\delta < 1$ such that $-\langle x, y\rangle \leq \delta \| x \|_2 \| y \|_2$, then
	\begin{equation}
	\| x + y \|_2 \geq \sqrt{\frac{1 - \delta}{2}} (\| x \|_2 + \| y
	\|_2)~.
	\end{equation}
\end{lemma}
\proof
	We know from \cite{mctr13} that
	\[ \| x + y \|_2^2 \geq (1 - \delta) (\| x \|_2^2 + \| y \|_2^2) \]
	and
	\[ (\| x \|_2 + \| y \|_2)^2 \leq 2 (\| x \|_2^2 + \| y \|_2^2) \]
	Combine them and we will get the conclusion.\qed

\subsection{Proof of Theorem \ref{prop:rho}}
\begin{theorem}[Structural Coherence (SC) Condition]
	Let $\delta := \max_i \delta_i$ with $\delta_i$ as defined in \myref{eq:deltai}. If $\delta<1$, there exists a $\rho>0$ such that for any $\Delta_i \in \cC_i, i = 1, \ldots,k$, the SC condition in \myref{cond2} holds, i.e.,
	\beq
	\left\| \sum_{i=1}^k \Delta_i \right\|_2 \geq \rho \sum_{i=1}^k \| \Delta_i \|_2~.
	\eeq
\end{theorem}
\proof
	We have by lemma (\ref{lemta})
	\[\textstyle
	 \left\| \sum_i \Delta_i \right\|_2 \geq \sqrt{\frac{1 - \delta}{2}}
	\left( \| \Delta_{i'} \|_2 + \left\| \sum_{j \neq i'} \Delta_k \right\|_2
	\right) . \]
	Sum over all possible combinations, we get
	\[\textstyle 
	k \left\| \sum_i \Delta_i \right\|_2 \geq \sqrt{\frac{1 -
			\delta}{2}} \sum_{i'} \left( \| \Delta_{i'} \|_2 + \left\| \sum_{j \neq
		i'} \Delta_k \right\|_2 \right) \geq  \sqrt{\frac{1 -
			\delta}{2}} \sum_{i'} \| \Delta_{i'} \|_2. \]
	Therefore
	\[\textstyle
	\left\|\sum_{i=1}^k\Delta_i\right\|_2 \geq \frac{1}{k} \sqrt{\frac{1 -
			\delta}{2}} \sum_{i=1}^k \|\Delta_i\|_2.
	\] \qed

\section{Restricted Eigenvalue Condition}
\subsection{Proof of Lemma \ref{lemcp}}\label{pf:lem:cp}
\begin{lemma}
	Let sets $\cC$ and $\cA$ be as defined in \myref{eq:setC} and \myref{eq:setA} respectively. If the SC condition in \myref{cond2} holds, then the marginal tail functions of the two sets have the following relationship:
	\beq
	Q_{\rho\xi}(\cH;Z) \geq Q_{\xi} (\cA ; Z).
	\eeq
\end{lemma}
\proof 	By definition, for any $u\in \cH$, we can find $u_i\in\cC_i, i=1,2,\ldots,k$ and $\sum_{i=1}^ku_i=u$. Then
\[\textstyle
|\langle Z,u\rangle| = \left\|\sum_{i=1}^k u_i\right\|_2
\left|\left\langle Z, \frac{u}{\|\sum_{i=1}^k u_i\|_2}
\right\rangle\right|\geq\rho\left|\left\langle Z,
\frac{u}{\|\sum_{i=1}^k u_i\|_2}\right\rangle\right|.
\]
Let $v = \frac{u}{\|\sum_{i=1}^k u_i\|_2}$, from definition
we know that $v\in A$. Hence we also have
\[\textstyle
P(|\langle Z, u\rangle|\geq \rho\xi)=
P\left(\frac{1}{\rho}|\langle Z, u \rangle|\geq \xi \right)
\geq P(|\langle Z, v\rangle|\geq \xi).
\]
Therefore taking the infimum over all $v\in \cA$ and then all $u
\in \cC$, the conclusion holds. \qed

\subsection{Proof of Theorem \ref{thm:re}}\label{pf:re}

\begin{theorem}[Restricted Eigenvalue Condition]
	Let $X$ be the sub-Gaussian design matrix that satisfies the assumptions above. If the SC condition \myref{cond2} holds with a $\rho > 0$, then with probability at least $1-\exp(-t^2/2)$, we have
	\beq
	\inf_{u \in \cH} \| X  u\|_2 \geq  c_1 \rho\sqrt{n} - c_2 w(\cH) - c_3 \rho t
	\eeq
	where $c_1, c_2$ and $c_3$ are positive constants determined by $\sigma_x$, $\sigma_\omega$ and $\alpha$.
\end{theorem}
\proof
Let two sets $\cC$ and $\cA$ be as defined previously. From Lemma (\ref{lemsb}) and lemma
(\ref{lemcp}) we know that for any $\xi>0$, with probability at least $1-e^{-t^2/2}$
\beq\label{bowl:ce}
  \inf_{u\in\cH}\|X u\|_2\geq \rho\xi\sqrt{n}
    Q_{2\xi}(\cA;Z)-2W(\cH;Z)-\rho\xi t.
\eeq
We use the "Bowling scheme" in \cite{trop15}, let $\mathbf{v}$ be any vector in $A$, by Paley-Zygmund inequality \cite{bolm13}, one can get
\beq\label{bowl:pz}
  P(|\lip x,v\rip|\geq 2\xi)\geq\frac{[E|
  \lip x,v\rip|-2\xi]_+^2}{E|\lip x,v\rip|^2} \geq \frac{(\alpha - 2\xi)^2}{4\sigma_x^2}.
\eeq

From the proof of \cite[Theorem~6.3]{trop15}, empirical width can be bounded by
\beq\label{bowl:gc}
W(\cH;Z) \leq L\sigma_x w(\cH)
\eeq

%
%

Select $\xi = \alpha/6$, combine \myref{bowl:ce}, \myref{bowl:pz}, \myref{bowl:gc} to discover that:
\[
  \inf_{u \in \cC} \| X u \|_2 \geq\frac{1}{9}\rho\alpha^3
  \sigma_x^{-2}\sqrt{n} - 2L\sigma_x w(\cH) -
  \rho\frac{\alpha}{6}t
\]
which completes the proof.\qed

From the conclusion above, the right hand side contains three parts. The first part is a constant times the square root of sample size, and the second part is a measure of the complexity of error sets. Therefore, when the number of samples is large enough or the error set has low complexity, the right terms will be larger than zero.

\subsection{Proof of Proposition \ref{prop:nec}}\label{pf:prop:nec}
\begin{proposition}
	If there is a matrix $X$ such that condition \eqref{eqre} holds for $\Delta_i \in \cC_i$, then SC \myref{cond2} holds.
\end{proposition}
\proof
If such $\rho$ does not exist, then there are some $\Delta_i \in \cC_i, i=1,\ldots,k$ not all zero such that
\[\textstyle
\left\| \sum_{i=1}^{k} \Delta_i \right\|_2 = 0 \Rightarrow \sum_{i=1}^{k} \Delta_i = 0,
\]
which implies $\|X\sum_{i=1}^{k} \Delta_i\|_2 = 0$ for every $X$. This is a contradiction. \qed

\section{Error Bound}

\subsection{Proof of Lemma \ref{lem:nd}}\label{pf:lem:nd}
\begin{lemma}
	Let design $X \in \R^{n \times p}$ be a row-wise i.i.d.~sub-Gaussian random matrix, and noise $\omega \in \R^n$
	be a centered sub-Gaussian random vector. 
	Then
	$
	s_n(\gamma) \leq c\frac{ w(\overline{\cH})}{\gamma \sqrt{n}}~.
	$
	for some constant $c>0$ with probability at least
	$1 - c_1\exp(-c_2 w^2(\overline{\cH})) - c_3\exp(-c_4n)$. Constant $c$ depends on $\sigma_x$ and $\sigma_\omega$.
\end{lemma}
\proof
	First notice that
	\[
	\omega^TX\Delta = \|\omega\|_2 . \frac{\omega X \Delta}{\|\omega\|_2}.
	\]
	We can first bound $\frac{\omega X \Delta}{\|\omega\|_2}$ then bound $\|\omega\|_2$.
	
	\noindent{\bf (a). Bound $\frac{\omega X \Delta}{\|\omega\|_2}$:}
	Note that $\frac{\omega X \Delta}{\|\omega\|_2}$ is not centered. In the first step we center it using
	\[\frac{1}{\|\omega\|_2}\epsilon_i\omega_ix_i^T\Delta,\]
	where $\epsilon_i$ is a Radmacher random variable, its probability of being $+1$ and $-1$ are both half. $x_i\in R^p$ is the i-th row of $X$. By assumption we know different $x_i$ are independent and have same distribution.
	
	Here we fix $\omega$.
	By proposition 5.10 in \cite{vers12}, the following bound holds:
	\beq
	P\left(\left|\frac{\omega X\Delta}{\|\omega\|_2}\right|\geq t\right)\leq e.\exp
	\left(-\frac{c_1t^2}{\sigma_x^2\|\Delta\|_2}\right).
	\eeq
	
	Through simple transform we know that
	\[
	P\left(\left|\left\lip\frac{\omega X}{\|\omega\|_2},u-v\right\rip
	\right|\geq t\right)\leq e.e^{-c_1 t^2/(\sigma_x^2
		\|u-v\|_2^2)}
	\]
	for any $u,v\in\mathbb{R}^p$. Then use \cite[Theorem~ 2.2.27]{tala14}
	\[
	P\left( \sup_{\Delta\in \overline{\cH}} \left|\frac{1}{\|\omega\|_2}\sum_i \epsilon_i \omega_i x_i^T\Delta \right| \geq c_2w(\overline{\cH}) + 2c_3t\right)
	\leq  c_4 \exp\left(-\frac{4 c_1 t^2}{\sigma_x^2}\right)
	\]
	
	\noindent{\bf (b). Bound $\omega$:} We first notice that $\|\omega\|_2^2$ is a sub-exponential
	random variable. Therefore if the sub-gaussian norm of $\omega$ is
	$\|\omega\|_{\phi_2}$, then for each entry $\omega_i$ the sub-expoential norm
	$\|\omega_i^2\|_{\phi_1}\leq 2\|\omega\|_{\phi_2}^2$. Through definition
	we reach:
	\[
	\mathbb{E}\|\omega\|_2\leq\sqrt{\mathbb{E}\|\omega\|_2^2}\leq\sigma_{\omega}\sqrt{2n}.
	\]
	Applying proposition 5.16 in \cite{vers12} to $\|\omega\|_2^2$, we obtain
	\[
	P(|\|\omega\|_2^2-\mathbb{E}\|\omega\|_2^2|\geq t)\leq
	2 \exp \left[-c_5 \min \left(\frac{t^2}{4\sigma_{\omega}^4},
	\frac{t}{2\sigma_{\omega}^2} \right)\right].
	\]
	Replace $t$ with $\sigma_{\omega}^2n$ gives
	\[
	P(\|\omega\|_2\geq 2\sigma_x\sqrt{n})\leq 2 \exp (-c_5 n).
	\]
	
	\noindent{\bf Combine (a) and (b):} First we have
	\[
	\begin{aligned}
	&P\left(\sup_{\Delta\in\overline{\cH}}\frac{1}{\sqrt{n}} \left|\sum_i\epsilon_i \omega_i x_i^T \Delta \right| \leq 2 c_2 \sigma_x w(\overline{\cH}) + 4 c_3 \sigma_x t \right)\\
	\geq & P\left(\sup_{\Delta\in \overline{\cH}} \left|\frac{1}{\|\omega\|_2}\sum_i \epsilon_i \omega_i x_i^T\Delta \right| \geq c_2w(\overline{\cH}) + 2c_3t \wedge \frac{\|\omega\|_2}{\sqrt{n}} \geq 2 \sigma_x \right)\\
	\geq & P\left(\sup_{\Delta\in \overline{\cH}} \left|\frac{1}{\|\omega\|_2}\sum_i \epsilon_i \omega_i x_i^T\Delta \right| \geq c_2w(\overline{\cH}) + 2c_3t |\omega \right) P\left( \frac{\|\omega\|_2}{\sqrt{n}} \geq 2 \sigma_x \right)\\
	\geq & 1 - 2\exp(-c_5 n) - c_4 \exp \left( - \frac{4 c_1 t^2}{\sigma_x^2}\right)
	\end{aligned}
	\]
	Then  choose $t = \frac{c_2}{2c_3}w(\overline{\cH})$,
	\[
	P\left(\sup_{\Delta\in\overline{\cH}}\frac{1}{\sqrt{n}}\left| \sum_i\epsilon_i\omega_i x_i^T\Delta \right| \leq 4 c_2 \sigma_{\omega}w(\overline{\cH})\right)
	\geq 1 - c_4.\exp\left(-\frac{c_1c_2^2w(\overline{\cH})}{c_3\sigma_x^2}\right) - 2\exp(-c_5n)
	\]
	
	Now we have bounded the symmetrized $\omega^TX\Delta$, then $\omega^TX\Delta$ can be bounded using symmetrization of probability \cite{leta13}:
	\[
	\begin{aligned}
	& P\left(\sup_{\Delta\in\overline{\cH}} \frac{1}{\sqrt{n}}|\omega^TX \Delta - E\omega^TX \Delta|> 16c_5\sigma_{\omega}w(\overline{\cH})\right) \\
	\leq & 4P\left(\sup_{\Delta\in\overline{\cH}} \frac{1}{\sqrt{n}}\left|\sum_i \epsilon_i \omega_i  x_i^T\Delta \right| > 4c_5\sigma_{\omega}w(\overline{\cH}) \right)
	\leq 4c_4.\exp\left(-\frac{c_1c_2^2w(\overline{\cH})}{c_3\sigma_x^2}\right) + 8\exp(-c_5n).
	\end{aligned}
	\]
	
	Because $\omega$ has zero mean, there the above inequality give us:
	\[
	\sup_{\Delta\in\overline{\cH}} \frac{|\omega^T X\Delta|}{\sqrt{n}} \leq 16 c_5 \sigma_{\omega} w(\overline{\cH})
	\]
	with high probability. Hence by definition
	\[
	s_n(\gamma) \leq \frac{16 c_5 \sigma_{\omega} w(\overline{\cH})}{\gamma \sqrt{n}}
	\]
	with probability at least $1 - 4c_4.\exp\left(-\frac{c_1c_2^2w(\overline{\cH})}{c_3\sigma_x^2}\right) - 8\exp(-c_5n)$. \qed

Now as we have the high probability bound of both $\kappa$ and $s_n(\gamma)$, we can derive our error bound for random case.


\subsection{Proof of Lemma \ref{lem:gw}}\label{pf:lem:gw}
\begin{lemma}[Gaussian width bound]
	Let $\cH$ and $\overline{\cH}$ be as defined in \myref{eq:setC} and \myref{eq:setBC} respectively. Then, we have
	$
	w (\cH) = O \left( \max_i w (\cC_i \cap \cS_{p-1}) + \sqrt{\log k} \right)
	$
	and
	$
	w(\overline{\cH}) = O \left( \max_i w (\cC_i \cap B_{p}) + \sqrt{\log k} \right)
	$.
\end{lemma}
\proof
By definition and the fact that the Gaussian width of convex hull of sets is equal to the Gaussian width of their union \cite{crpw12}
\[ w (\cH) =\mathbbm{E} \sup_{u \in \cH} \langle u, g \rangle =\mathbbm{E}
\max_i \sup_{u_i \in \cC_i\cap S_{p-1}} \langle u_i, g \rangle \]
By concentration inequality for Lipschitz functions \cite{leta13}, for each $i=1,\ldots, k$
\[ P (\sup_{u_i \in \cC_i\cap S_{p-1}} \langle u_i, g \rangle \geq \mathbbm{E}
\sup_{u_i \in \cC_i\cap S_{p-1}} \langle u_i, g \rangle + r) \leq \exp ( -
r^2 /2 ) . \]
Then denote $D_i = \sup_{u_i \in \cC_i\cap S_{p-1}} \langle u_i, g \rangle$, we have
\[  \begin{aligned}
w(\cH) =&  \mathbbm{E} \max_i D_i\\
\leq & \max_i \mathbbm{E} D_i + \delta + \sum_{i=1}^k \int_{\delta}^{\infty} P \left(D_i \geq \max_i \mathbbm{E} D_i + r\right) d r\\
\leq & \max_i \mathbbm{E} D_i + \delta + k
\int_{\delta}^{\infty} \exp ( - r^2/2 ) d r\\
\leq & \max_i \mathbbm{E} D_i + \delta + \eta' k \exp (-
\delta^2 / 2) .
\end{aligned} \]
Let $\delta = \sqrt{\log k}$, we get
\[
w (\cH) \leq \max_i \mathbbm{E} D_i +
\sqrt{\log k} + \eta = O \left( \max_i w (\cC_i \cap S_{p - 1}) +\sqrt{\log k} \right).
\]
the conclusion holds. The conclusion for $w(\overline{\cH})$ can be proved the same as above.
\qed

\subsection{Proof of Theorem \ref{th:bnd1}}
\begin{theorem}
	For estimator \myref{eq:bnd0}, let $\cC_i=\cone \{ \Delta: R_i(\theta_i^*+\Delta) \leq R_i(\theta_i^*) \}$, design $X$ be a random matrix with each row an independent copy of sub-Gaussian random vector $Z$, noise $\omega$ be a centered sub-Gaussian random vector, and $B_p \subseteq\mathbb{R}^p$ be the a centered unit euclidean ball. If sample size $n > c (\max_i w^2(\cC_i\cap S_{p-1}) + \log k )/\rho^2$, then with high probability,
	\beq
	\sum_{i=1}^k \| \hat{\theta_i}-\theta_i^* \|_2 \leq C \frac{\max_i w (\cC_i \cap B_{p}) + \sqrt{\log k}}{\rho^2\sqrt{n}},
	\eeq
	for constants $c, C>0$ that depend on sub-Gaussian norms $\vertiii{Z}_{\phi_2}$ and $\vertiii{\omega}_{\phi_2}$.
\end{theorem}
\proof
  Firstly, we choose
  \[
  t = \frac{1}{3}\alpha^2\sigma_x^{-2} \sqrt{n} - 6L\sigma_x\rho^{-1} \alpha^{-1} w(\cH).
  \]
  From theorem \ref{thm:re}, RE condition holds for
  \[
  \kappa \geq \frac{1}{2}\left(\frac{1}{9}\rho\alpha^3 \sigma_x^{-2} - 2 L \sigma_x \frac{w(\cH)}{\sqrt{n}}\right)
  \]
  with probability at least
  \[
  1 - \exp\left(-(\frac{1}{3}\alpha^2\sigma_x^{-2} \sqrt{n} - 6L\sigma_x\rho^{-1} \alpha^{-1} w(\cH))^2/2\right).
  \]
	
  Next we choose $\gamma = \frac{\rho^2\alpha^6}{1296\sigma_x^4}$, and let $s_n(\gamma)$ be defined as above.
  Thus from theorem (\ref{thdet}) and our discussion above, if
  \[
    \frac{1}{4}\left(\frac{1}{9}\rho\alpha^3 \sigma_x^{-2} - 2 L \sigma_x \frac{w(\cH)}{\sqrt{n}}\right)^2 >  \frac{\rho^2\alpha^6}{1296\sigma_x^4} \Rightarrow n > c_1 w^2(\cH),
  \]
  for some constant $c_1>0$, then $\kappa > 2\gamma$. Using theorem \ref{thdet}, we have
  \[
   \sum_{i=1}^k \|\Delta_i\|_2 \leq c_2\frac{w(\overline{\cH})}{\sqrt{n}}.
  \]
  with probability at least
  \[
  1 - c_3\exp(-c_4w(\overline{\cH})) - c_5\exp(-c_6n) - \exp(-(c_7\sqrt{n}-c_8w(\cH))^2),
  \]
  which completes the proof.\qed

\section{Examples}\label{pf:ex}

\subsection{Structural Coherence For 1-sparse + 1-sparse MCA}\label{pf:sc11}
\begin{proposition}
	Suppose both vector $\theta_1$ and vector $Q\theta_2$ are one sparse, and the i-th entry of $\theta_1$ and the j-th entry of $Q\theta_2$ are non-zero. If 
	\[
	Q_{ij}\sign(\theta_1^*)_i\sign(Q\theta_2^*)>0,
	\]
	then we have
	\beq
	\rho \geq \sqrt{(1 - \sqrt{1 - Q_{ij}^2})/2}.
	\eeq
\end{proposition}
\proof
	Denote $\Delta_{-i}$ as a vector whose ith entry is 0, and other entries are equal to those of $\Delta$. Suppose both $\theta_1$ and $Q\theta_2$ are 1-sparse vectors, and the ith entry of $\theta_1$ jth entry of $Q\theta_2$ are nonzero. Then error vector $\Delta_1$ and $\Delta_2$ satisfy the following inequalities:
	\[
	-\langle\sign(\theta_{1i}), \Delta_{1i}\rangle \geq \|\Delta_{1-i}\|_1, -\langle\sign(\theta_{2j}), \Delta_{2j}\rangle \geq \|\Delta_{2-j}\|_1.
	\]
	Therefore
	\[
	\frac{-\langle\sign(\theta_{1i}), \Delta_{1i}\rangle}{\|\Delta_1\|_2}\geq \frac{1}{\sqrt{1 + \frac{\|\Delta_{1-i}\|_2^2}{\Delta_{1i}^2}}} \geq \frac{1}{ \sqrt{1 +  \frac{\|\Delta_{1-i}\|_2^2}{\|\Delta_{1-i}\|_1^2}}} \geq \frac{1}{\sqrt{2}}.
	\]
	The same holds for $Q\Delta_2$.
	Then when $Q_{ij}\sign(\theta_{1i})\sign(Q\theta_{2j})>0$ we have from geometry that
	\[
	-\langle \Delta_1, \Delta_2 \rangle \leq -\cos(2\arccos (\frac{1}{\sqrt{2}})+ \arccos(Q_{ij}\sign(\theta_{1i})\sign(Q\theta_{2j}))) \leq \sqrt{1 - Q_{ij}^2}
	\]
	Therefore $\rho \geq \sqrt{\frac{1-\sqrt{1-Q_{ij}^2}}{2}}$ by proposition \ref{lemta}.
\qed

\subsection{Proof of Theorem \ref{th:mcal1}}
\begin{theorem}
	If $M \leq  \frac{1}{8\sqrt{s_1s_2}}$, then for problem \eqref{lle} with high probability
	\[
	\|\theta_1 - \theta_1^*\|_2 + \|\theta_2 - \theta_2^*\|_2  = O\left( \max\left\{\sqrt{\frac{s_1\log p}{n}}, \sqrt{\frac{s_2\log p}{n}}\right\} \right).
	\]
\end{theorem}
\proof
Let $\Delta_i = \theta_i - \theta_i^*$ for $i=1,2$, then we have
\[
\begin{aligned}
\langle\Delta_1, \Delta_2\rangle & = \langle\Delta_1, Q^TQ\Delta_2\rangle \leq \max_{ij}|Q_{ij}|\|\Delta_1\|_1\|Q\Delta_2\|_1 \\
 & \leq M(\|P_{\Omega}\Delta_1\|_1 + \|P_{\Omega_1^c}\Delta_1\|_1) (\|P_{\Omega_2}Q\Delta_2\|_1 + \|P_{\Omega_2^c}Q\Delta_2\|_1)\\
 & \leq 4M\|P_{\Omega}\Delta_1\|_1\|P_{\Omega_2}Q\Delta_2\|_1 \leq 4M\sqrt{s_1s_2}\|\Delta_1\|_2\|Q\Delta_2\|_2
\end{aligned}
\]
Let $4M\sqrt{s_1s_2} \leq \frac{1}{2}$, we get $M \leq \frac{1}{8\sqrt{s_1s_2}}$ and $\rho \geq \frac{1}{2}$.

From the result of \cite{crpw12}, we know that the Gaussian width for the error cone of a $s$-sparse vector is $O (\sqrt{s\log (\frac{p}{s}) })$. Therefore by theorem \ref{th:bnd1}, if $n\geq C\max_{s\in\{s_1, s_2\}} s\log (\frac{p}{s})$ for some $C>0$, then
\[
	\|\theta_1 - \theta_1^*\|_2 + \|\theta_2 - \theta_2^*\|_2  = O\left( \max\left\{\sqrt{\frac{s_1\log p}{n}}, \sqrt{\frac{s_2\log p}{n}}\right\} \right).
\]
\qed

\subsection{Proof of Theorem \ref{th:mcakk}}
\begin{theorem}
	If $\sigma \leq \frac{1}{4(1+\frac{\theta_{1max}}{\theta_{1min}})(1+\frac{\theta_{2max}}{\theta_{2min}})}$ where $\theta_{max} = \max_{i\in \supp(\theta)}|\theta_i|$ and $\theta_{min} = \min_{i\in \supp(\theta)}|\theta_i|$, then we have for problem \eqref{kke} with high probability
	\[
	\|\theta_1 - \theta_1^*\|_2 + \|\theta_2 - \theta_2^*\|_2 = O\left(\max\left\{\sqrt{\frac{s_1\log(p-s_1)}{n}},\sqrt{\frac{s_2\log(p-s_2)}{n}}\right\}\right).
	\]
\end{theorem}
\proof
	We first characterize the interaction between cones:
	\[
	\langle\Delta_1, \Delta_2\rangle  = \langle\Delta_1, Q^TQ\Delta_2\rangle.
	\]
	because $k$-support norm is an atomic norm, and its atomic set is all unit $k$-sparse vectors. Therefore we can decompose $\Delta_1$ into combination of unit $s_1$-sparse vectors $\Delta_1 = \sum_i \alpha_iu_i$ and $Q\Delta_2$ into combination of unit $s_2$-sparse vectors $Q\Delta_2 = \sum_j \beta_jv_j$. Then
	\[
	\langle\Delta_1, Q^TQ\Delta_2\rangle \leq \sum_i|\alpha_i|\sum_j|\beta_j|\max_{ij}|u_i^TQ^Tv_j|\leq \sigma\|\Delta_1\|_{s_1}^{sp}\|\Delta_2\|_{s_2}^{sp}.
	\]
	By Theorem 9 in \cite{chba15}, we have 
	\[
	\|\Delta_1\|_{s_1}^{sp} \leq \sqrt{2}(1+\frac{\theta_{1max}}{\theta_{1min}})\|\Delta_1\|_2\quad \text{and}\quad \|\Delta_2\|_{s_2}^{sp} \leq \sqrt{2}(1+\frac{\theta_{2max}}{\theta_{2min}})\|\Delta_1\|_2.
	\]
	Therefore
	\[
	\langle\Delta_1, \Delta_2\rangle \leq 2\sigma(1+\frac{\theta_{1max}}{\theta_{1min}}) (1+\frac{\theta_{2max}}{\theta_{2min}}) \|\Delta_1\|_2\|\Delta_2\|_2.
	\]
	Let $2\sigma(1+\frac{\theta_{1max}}{\theta_{1min}}) (1+\frac{\theta_{2max}}{\theta_{2min}}) \leq \frac{1}{2}$,
	then $\sigma \leq \frac{1}{4(1+\frac{\theta_{1max}}{\theta_{1min}}) (1+\frac{\theta_{2max}}{\theta_{2min}})}$
	and $\rho \geq \frac{1}{2}$.
	
	From \cite{chba15}, when we set $k$ to be the sparsity $s$, the corresponding Gaussian width of tangent cone is $O(s\log(p-s))$, therefore plus this result in Theorem \ref{th:bnd1} we get
	\[
	\|\theta_1 - \theta_1^*\|_2 + \|\theta_2 - \theta_2^*\|_2 = O\left(\max\left\{\sqrt{\frac{s_1\log(p-s_1)}{n}},\sqrt{\frac{s_2\log(p-s_2)}{n}}\right\}\right).
	\]
\qed

\subsection{Proof of Theorem \ref{th:rpca}}
\begin{theorem}
	If there is a $\rho>0$ for problem \eqref{lsd}, then with high probability
	\[
	\|L - L^*\|_2 + \|S - S^*\|_2 = O\left(\max\left\{\sqrt{\frac{s\log(d_1d_2-s)}{n}}, \sqrt{\frac{r(d_1+d_2-r)}{n}}\right\} \right).
	\]
\end{theorem}
\proof
	From \cite{crpw12}, we know that for error cone of $d_1\times d_2$ rank-r matrix, its Gaussian width is $O(r(d_1 + d_2 - r))$. Therefore if $\rho >0$, then by applying theorem \ref{th:bnd1}, the error bound is
	\[
		\|L - L^*\|_2 + \|S - S^*\|_2 = O\left(\max\left\{\sqrt{\frac{s\log(d_1d_2-s)}{n}}, \sqrt{\frac{r(d_1+d_2-r)}{n}}\right\} \right).
	\]
\qed

\subsection{Additional Example for Low-rank ans Sparse Matrix Decomposition}\label{ex:rpca11}
\begin{example}
	Suppose noise $\omega=0$,
	\beq
	S_0 = L_0 = \left(\begin{matrix}
		1 & 0 & \ldots & 0\\
		0 & 0 & \ldots & 0\\
		\vdots & \vdots & \ddots & \vdots \\
		0 & 0 & \ldots & 0
	\end{matrix}\right),\label{slm}
	\eeq
	and $M = S_0 + L_0$, then the SC condition of problem \myref{lsd} holds.
\end{example}

\proof
	Suppose the singular value decomposition of $L_0^*$ is $U\Sigma V^T$. Denote
	\[
	\cC_S'=cl\{\Delta|-\langle sign(S_0'),\Delta\rangle \geq \|P_{\Omega^c}(\Delta)\|_1\},~
	\cC_L'=cl\{\Delta|-\langle U V^T, \Delta\rangle \geq \|P_{T^\perp}(\Delta)\|_*\}.
	\]
	From \cite{mctr13} we know that SC is equivalent to $-\cC_S'\bigcap\cC_L' = \{0\}$. To prove  $-\cC_S'\bigcap\cC_L' = \{0\}$, we need the following inequalities:
	\[
	\langle sign(S_0'),\Delta\rangle \geq \|P_{\Omega^c}(\Delta)\|_1,~
	-\langle U V^T, \Delta\rangle \geq \|P_{T^\perp}(\Delta)\|_*.
	\]
	has unique solution 0.
	
	It is easy to notice that $\langle sign(S_0'),\Delta\rangle = \langle U V^T, \Delta\rangle = \Delta_{11}$. As the value of norms is non-negative, we have
	$
	\Delta_{11} \geq 0
	$
	and
	$
	-\Delta_{11} \geq 0
	$.
	Therefore $\Delta_{11}=0$. Besides,
	\[
	\textstyle \|P_{\Omega^c}(\Delta)\|_1 = \sum_{(i,j)\neq (1,1)} |\Delta_{ij}| \leq 0,
	\]
	which leads to $\Delta_{ij} = 0$ for $(i,j)\neq (1,1)$.
	
	Finally, $\Delta=0$ and the conclusion holds.\qed

People tend to think that we cannot obtain the correct decomposition in this situation.
Note that the cone $\cC_S$ is centered at one point, and the cone $\cC_L$ contains $\cC_S$ but their surface contacts only at the origin. 
Therefore the reflection of one cone will touch the other cone only at the origin. As a result, for $M = M_0 = S_0+L_0$, i.e., SC condition holds.

\section{Noiseless Case: Comparing Estimators}

In this section we try to explore the structures that are different between problem \myref{eq:est1} and problem \myref{eq:est2} and the structures that they share.

\subsection{Proof of Lemma \ref{lemic}}\label{pf:lemic}
\begin{lemma}
	For a given set $\{\lambda_i\}$, the infimal convolution norm ball $\Omega_R$ is the convex hull of
	$\bigcup_{i=1}^k \frac{1}{\lambda_i}\Omega_R^i$, i.e., $\Omega_R = \conv(\bigcup_{i=1}^k \frac{1}{\lambda_i}\Omega_R^i)$.
\end{lemma}
\proof
	If $\theta \in \Omega_R$, then from definition there are $\sum_{i=1}^k\theta_i=\theta$
	such that $R(\theta) = \sum_{i=1}^k \lambda_i R_i(\theta_i)$.
	Without loss of generalization, suppose $\theta_i \neq 0$ for each $i$, then we have
	the following decomposition:
	\begin{equation}\label{esdec}
	\sum_{i=1}^k \frac{\lambda_iR_i (\theta_i)}{R(\theta)}
	\frac{R(\theta)}{\lambda_i R_i (\theta_i)} \theta_i= \theta.
	\end{equation}
	It is easy to know that $\sum_{i=1}^k \frac{\lambda_iR_i (\theta_i)}{R(\theta)} = 1$
	and $R_i(\frac{R(\theta)}{\lambda_i R_i (\theta_i)} \theta_i) = \frac{1}{\lambda_i}
	R(\theta) \leq \frac{1}{\lambda_i}$.
	Therefore $\frac{R(\theta)}{\lambda_i R_i (\theta_i)} \theta_i \in
	\frac{1}{\lambda_i} \Omega_R^i$, and $\theta \in conv(\bigcup_{i=1}^k \frac{1}{\lambda_i}\Omega_R^i)$.
	
	If $\theta \in conv(\bigcup_{i=1}^k \frac{1}{\lambda_i}\Omega_R^i)$, then we can find $\theta_i
	\in \frac{1}{\lambda_i} \Omega_R^i$ and $c_i>0, \sum_{i=1}^kc_i = 1$ such that
	\[\sum_{i=1}^k c_i \theta_i = \theta.\]
	Then
	\[
	R(\theta)\leq\sum_{i=1}^k \lambda_i R_i(c_i \theta_i)= \sum_{i=1}^k
	c_i \lambda_i R_i(\theta_i) \leq 1.
	\]
	Therefore $\theta \in \Omega_R$ which completes the proof.\qed

\subsection{Proof of theorem \ref{thpa}}\label{pf:thpa}
\begin{theorem}
	Given $\hat{\theta}_1, \ldots, \hat{\theta}_k$ and define
	\beq\label{def:c0d}
	\cC_0 = \left\{\sum_{\theta_i\neq 0} \left(\frac{c_i'}{c_i}-1\right)\theta_i~|~c_i'\geq 0, \sum_{i=1}^k c_i'=1\right\}.
	\eeq
	Suppose $dim(span\{\theta_i\})=k$,
	then there exist $\lambda_1, \ldots, \lambda_k$ such that $\sum_{i=1}^k \hat{\theta}_i = \theta$ are unique solutions of (\ref{defic}) if and only if there are $c_1,\ldots,c_k$ with $c_i\geq 0$ and $\sum_{i=1}^kc_i = 1$ such that for the corresponding error cone $\cC_i$ of $\hat{\theta}_i$ and $C_0$ defined above,
	$
	- \cC_i\cap \sum_{j \neq i} \cC_j = \{ 0 \},
	$
	for $i=0,1,\ldots,k$.
\end{theorem}
\proof
Before proofing the main result we need the following lemma, it is proved in appendix \ref{pf:lem:fix}.
	\begin{lemma}
		\label{lem:fix}
		For fixed $\lambda_1, \ldots, \lambda_k$, suppose $\sum_{i=1}^k\theta_i = \theta$ is a solution of decomposition (\ref{defic}) under this set of $\{\lambda_i\}$, and $dim(span\{\theta_i\})=k$. Let $\cC_i,i=1,2,\ldots,k$ be the corresponding error cones of $\{\theta_i\}$, $c_i = \lambda_iR_i(\theta_i) / R(\theta)$ and $\cC_0$ be as defined in \eqref{def:c0d}.
		The decomposition (\ref{defic}) for $\theta$ is unique if and only if for any
		$i=0,1,\ldots,k$,
		\beq
		- \cC_i\cap \sum_{j \neq i} \cC_j = \{ 0 \}.
		\eeq
	\end{lemma}

	We come to the main result and the necessity is obvious from lemma \ref{lem:fix};
	
	Without loss of generality, suppose $c_i\geq 0$. If such $c_1,\ldots,c_k$ exist for $\hat{\theta}_1,\ldots,\hat{\theta}_k$, let $\lambda_i = \frac{c_i}{R_i(\hat{\theta}_i)}$.
	Suppose $\theta_1,\ldots,\theta_k$ is a set of optimal solution under the $\lambda$ defined above. From \ref{esdec} we can write the decomposition as $\theta = \sum_{i=1}^k c_i'\theta_i'$ where $c_i'$ is a coefficient of convex combination, $c_i'\theta_i' = \theta_i$ and $\lambda_i R_i(\theta_i') = R(\theta)$ under coefficient $\lambda_i$.
	
	If $\theta_i \neq \hat{\theta}_i$ for some i, then as $\sum_{i=1}^k\lambda_iR_i(\hat{\theta}_i) = \sum_{i=1}^k c_i = 1$, we have
	\[
	\lambda_iR_i(\theta_i)\leq 1 \Rightarrow R_i(c_i\theta_i)\leq R_i(\hat{\theta}_i).
	\]
	Therefore $c_i\theta_i - \hat{\theta}_i \in \cC_i$ by definition.
	We also have
	\[
	\sum_{i=1}^k\hat{\theta}_i + \sum_{i=1}^k(\frac{c_i'}{c_i}-1)\hat{\theta}_i + \sum_{i=1}^k \frac{c_i'}{c_i}(c_i\theta_i - \theta_i)  = \theta \Rightarrow \frac{c_i'}{c_i}(c_i\theta_i - \theta_i) = -\sum_{i=1}^k(\frac{c_i'}{c_i}-1)\hat{\theta}_i
	\]
	which is contradict to our condition. Therefore $\hat{\theta}_i$ is a solution. Uniqueness is a direct conclusion of lemma \ref{lem:fix}.\qed

Note that in this proof we set $\lambda_i = \frac{c_i}{R_i(\theta_i)}$. This is also a general way to choose the parameter $\{\lambda_i\}$.

\subsection{Proof of Lemma \ref{lem:fix}}\label{pf:lem:fix}

\proof
	\
	Without loss of generality, assume $\theta_i\neq 0$.According to (\ref{esdec}),
	let
	\[c_i = \frac{\lambda_i R_i(\theta_i)}{R(\theta)}, and~~\theta_i' =
	\frac{1}{c_i}\theta_i.\]
	Suppose $\theta_i$ is a unique decomposition, for any $\Delta_i \in \cC_i$,
	if $ \sum_{i=1}^k c_i' \lambda_i R_i (\theta_i' + \Delta_i) \leq R (\theta) $ and
	$\sum_{i=1}^k c_i'(\theta_i'+ \Delta_i) = \theta$ for some $c_i'\geq 0,
	\sum_{i=1}^k c_i' = 1$. Therefore we obtain the following decomposition of $\theta$:
	\[
	\theta = \sum_{i=1}^k c_i\theta_i' + \sum_{i=1}^k (c_i'-c_i)\theta_i +\sum_{i=1}^k
	c_i'\Delta_i.
	\]
	It is obvious from observation that $\sum_{i=1}^k (c_i'-c_i)\theta_i +\sum_{i=1}^k
	c_i'\Delta_i = 0$ and $\sum_{i=1}^k (c_i'-c_i)\theta_i \in \cC_0$, $\sum_{i=1}^k
	c_i'\Delta_i \in \sum_{i=1}^k \cC_i$.
	
	By minimal of $R(\theta)$, $\sum_{i=1}^k c_i' \lambda_i R_i (\theta_i' + \Delta_i) =
	R (\theta)$. By our assumption, such decomposition of $\theta$ is unique, thus
	\[
	\sum_{i=1}^k (c_i'-c_i)\theta_i = \sum_{i=1}^k c_i'\Delta_i = 0.
	\]
	which implies $-\cC_0\cap \sum_{i=1}^k \cC_i=\{0\}$. Uniqueness also give that
	$c_i'\Delta_i = 0$ for each $i=1,2,\ldots,k$. Therefore $- \cC_i\cap \sum_{j \neq i}
	\cC_j = \{ 0 \}$.
	
	If $\theta_i$ is not a unique decomposition then there are some $\Delta_i \neq
	0$ such that $\sum_{i=1}^k \Delta_i = 0$ and
	\[
	\sum_{i=1}^k \lambda_i R_i (\theta_i + \Delta_i) = R (\theta).
	\]
	Let
	\[c_i'' = \frac{\lambda_i R_i(\theta_i+\Delta_i)}{R(\theta)}, and~~\theta_i'' =
	\frac{1}{c_i''}(\theta_i+\Delta_i),\]
	we have
	\[ \lambda_i R_i (\theta_i'') = R (\theta), \]
	and hence $\theta_i'' - \theta_i' \in \cC_i$ for $\lambda_i R_i(\theta_i')=R(\theta)$.
	Unfold $\theta_i'$ and $\theta_i''$ gives
	\[
	c_i''(\theta_i''-\theta_i')=c_i''(\frac{1}{c_i''}-\frac{1}{c_i'})\theta_i +\Delta_i.
	\]
	Sum over all $i$, we get
	\[
	\sum_{i=1}^k c_i''(\theta_i''-\theta_i')=\sum_{i=1}^k c_i''(\frac{1}{c_i''}-
	\frac{1}{c_i'})\theta_i,
	\]
	which is contradict to our assumption that $-\cC_0\cap \sum_{i=1}^k \cC_i=\{0\}$.
	Therefore the conclusion holds.\qed

\end{document}